%% file: ms.tex
\documentclass[runningheads]{llncs}

\usepackage{eccv}

\input{preamble}

\usepackage{eccvabbrv}

\usepackage{graphicx}
\usepackage{booktabs}
\usepackage{wrapfig}

\usepackage[accsupp]{axessibility}  

\usepackage[pagebackref,breaklinks,colorlinks,citecolor=eccvblue]{hyperref}

\usepackage{orcidlink}

\newcommand{\methodname}{PartCraft}

\begin{document}

\title{PartCraft: Crafting Creative Objects by Parts}

\titlerunning{PartCraft: Crafting Creative Objects by Parts}

\author{Kam Woh Ng\inst{1,2}\orcidlink{0000-0002-9309-563X} \and
Xiatian Zhu\inst{1,3}\orcidlink{0000-0002-9284-2955} \and
Yi-Zhe Song\inst{1,2}\orcidlink{0000-0001-5908-3275} \and
Tao Xiang\inst{1,2}}

\authorrunning{KW. Ng et al.}

\institute{CVSSP, University of Surrey, United Kingdom \and
iFlyTek-Surrey Joint Research Centre \and
Surrey Institute for People-Centred AI\\
\email{\{kamwoh.ng,xiatian.zhu,y.song,t.xiang\}@surrey.ac.uk}}

\maketitle

\input{sec/0_abstract}    
\input{sec/1_intro} 
\input{sec/2_related_work}
\input{sec/3_methodology}
\input{sec/4_experiment}
\input{sec/5_conclusion}

%
%
\bibliographystyle{splncs04}
\bibliography{main}

\input{sec/X_suppl}

\end{document}

%% file: preamble.tex
\usepackage[dvipsnames]{xcolor}

\usepackage{adjustbox}
\usepackage{multirow}
\usepackage{caption}
\usepackage{subcaption}
\usepackage{amssymb}
\usepackage{pifont}
\usepackage[dvipsnames]{xcolor}
\newcommand{\cmark}{\textcolor{Green}{\text{\ding{51}}}}%
\newcommand{\xmark}{\textcolor{Red}{\text{\ding{55}}}}%

\usepackage{amsmath}

\DeclareMathOperator*{\argmin}{argmin}

\usepackage{algorithm}
\usepackage{algpseudocode}

\usepackage{newfloat}
\usepackage{listings}
\DeclareCaptionStyle{ruled}{labelfont=normalfont,labelsep=colon,strut=off} 
\floatstyle{ruled}
\newfloat{listing}{tb}{lst}{}
\floatname{listing}{Listing}

\usepackage{etoolbox}
\makeatletter
\AfterEndEnvironment{algorithm}{\let\@algcomment\relax}
\AtEndEnvironment{algorithm}{\kern2pt\hrule\relax\vskip3pt\@algcomment}
\let\@algcomment\relax
\newcommand\algcomment[1]{\def\@algcomment{\footnotesize#1}}
\renewcommand\fs@ruled{\def\@fs@cfont{\bfseries}\let\@fs@capt\floatc@ruled
  \def\@fs@pre{\hrule height.8pt depth0pt \kern2pt}%
  \def\@fs@post{}%
  \def\@fs@mid{\kern2pt\hrule\kern2pt}%
  \let\@fs@iftopcapt\iftrue}
\makeatother

\usepackage{float}

%% file: sec/0_abstract.tex
\begin{abstract}

This paper propels creative control in generative visual AI by allowing users to ``select''. Departing from traditional text or sketch-based methods, we for the first time allow users to choose visual concepts by parts for their creative endeavors. The outcome is fine-grained generation that precisely captures selected visual concepts, ensuring a holistically faithful and plausible result. 
To achieve this, we first parse objects into parts through unsupervised feature clustering.
Then, we encode parts into text tokens and introduce an entropy-based normalized attention loss that operates on them. This loss design enables our model to learn generic prior topology knowledge about object's part composition, and further generalize to novel part compositions to ensure the generation looks holistically faithful. 
Lastly, we employ a bottleneck encoder to project the part tokens. This not only enhances fidelity but also accelerates learning, by leveraging shared knowledge and facilitating information exchange among instances. 
Visual results in the paper and supplementary material showcase the compelling power of \methodname{} in crafting highly customized, innovative creations, exemplified by the ``charming'' and creative birds in Fig.~\ref{fig:big-teaser}. Code is released at \url{https://github.com/kamwoh/partcraft}.

\keywords{Part Composition \and Controllable Text-to-image Generation} 

\end{abstract}

%% file: sec/1_intro.tex
\section{Introduction}
\label{sec:intro}

\begin{figure*}[t]
    \centering
    \includegraphics[width=\textwidth]{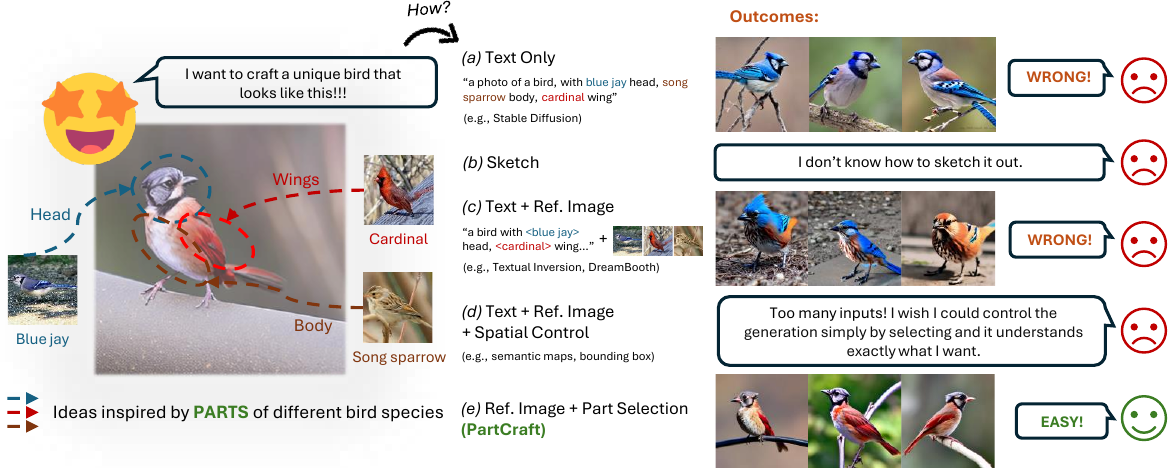}
    \caption{
    People often form creative concepts based on existing ones \cite{wilkenfeld2001similarity, nagai2009concept, runco2012standard, bonnardel2005towards}. For instance, a bird enthusiast may want to craft a unique bird with different parts (\eg, heads, bodies, and wings) from common bird types (\eg, blue jay, cardinal and song sparrow).
    \textbf{(a)} Using text prompts in T2I models often results in a lack of control and deviation from the intended details, especially those visual details that are difficult to describe.
    \textbf{(b)} While sketching is a direct way, not everyone possesses the ability to sketch, particularly in intricate detail.
    \textbf{(c)} Even with reference images, existing methods (\eg, DreamBooth \cite{ruiz2023dreambooth}) did not consider learning object parts, thus cannot generate with desired parts.
    \textbf{(d)} Using additional control is even cumbersome, requiring too many inputs!
    \textbf{(e)} We aim to create an object by simply selecting desired parts. \methodname{} learns from visual examples to generate the object with a faithful holistic structure, seamlessly integrating the chosen parts into a natural and coherent entity.
    }
    \label{fig:big-teaser}
    \vspace{-0.4cm}
\end{figure*}

Humans are creators; AI, on the other hand, hallucinates. Creativity, arguably, is the very force driving humanity forward. Recently, generative AI has garnered considerable attention for its perceived ``creativity'' \cite{rombach2022ldm_sd, ramesh2022hierarchical_dalle, balaji2022ediffi, ding2022cogview2, nichol2021glide, saharia2022photorealistic_imagen, ding2021cogview, qu2023sketchdreamer, qu2024wired, han2024headsculpt}. Despite its promise, the challenge of control has swiftly surfaced -- how can humans infuse their creativity into the generation process and regulate the extent to which AI hallucinates?

Creativity starts with an idea. The immediate challenge is how to articulate that idea and integrate it with generative AI. Text is the most commonly employed medium. For instance, imagine being a bird enthusiast wanting to craft the most unique bird akin to Fig.~\ref{fig:big-teaser}. The go-to approach would be furnishing Stable Diffusion \cite{rombach2022ldm_sd} (or an equivalent model) with the following textual prompt (the idea): \textit{``generate a bird with the head of X, wings of Y, and body of Z''.} While you might be presented with remarkably looking birds like Fig.~\ref{fig:big-teaser}(a), they may bear little resemblance to the envisioned concept. Recent literature suggests that a swift sketch could serve as a viable alternative \cite{mo2023freecontrol, zhang2023controlnet, wang2024instancediffusion}, providing fine-grained shape control. However, the caveat is that not everyone possesses the ability to sketch, particularly in intricate detail.

In this paper, our primary focus revolves around addressing the issue of ``control'' in generative AI. We advance by introducing fine-grained control into the generative process, inviting you to ``select''. While this selection mechanism might seem modest at first, it closely mirrors the human creative process, where new concepts often emerge from existing ones \cite{wilkenfeld2001similarity, nagai2009concept, runco2012standard, bonnardel2005towards}. Recall those moments when you desired an ``ideal'' pair of shoes with selected features from different pairs, or when you aimed to get creative with a cat (for that matter!)?

\begin{figure*}[t]
    \centering
    \includegraphics[width=\textwidth]{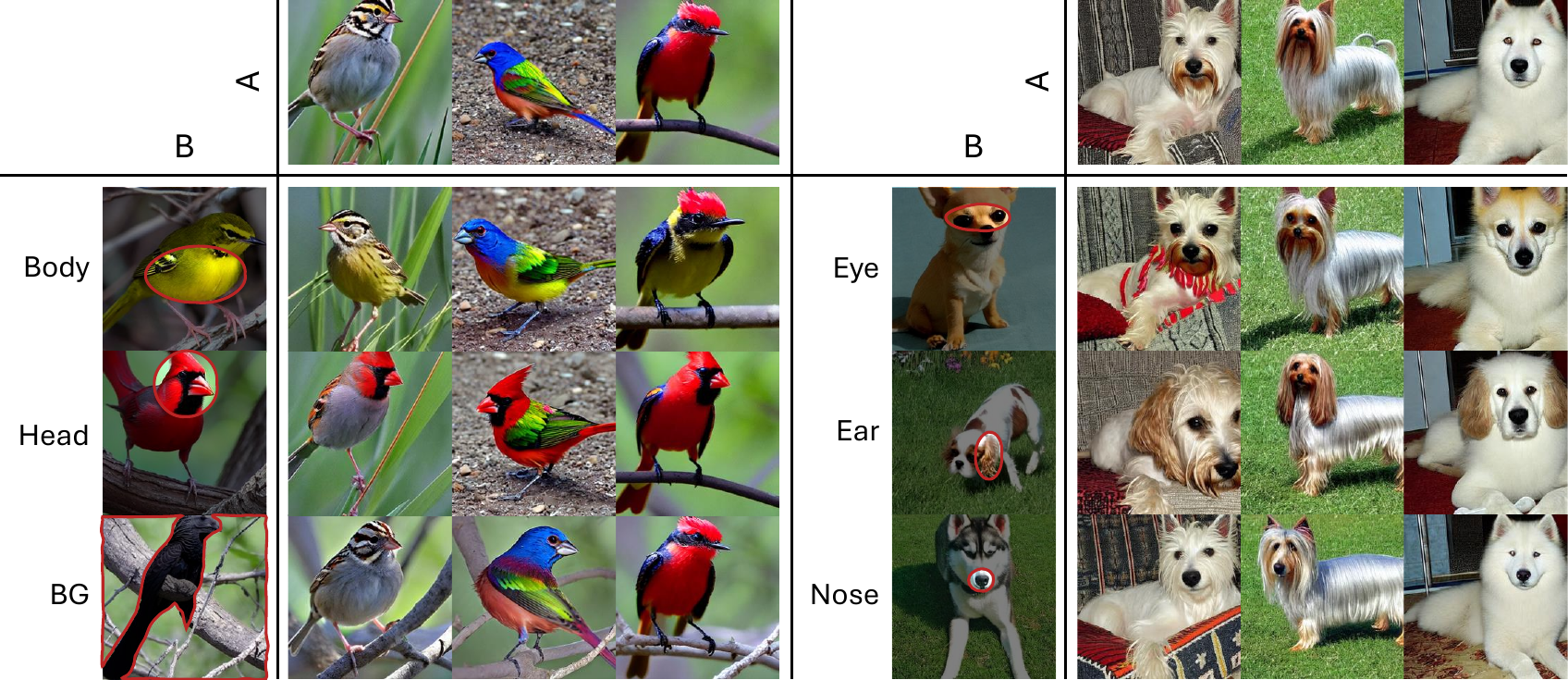}
    \caption{
    Two sets of images were generated from their original parts (sources A and B). We can integrate a specific part (\eg, body, head, or even background (BG) of source B to target A seamlessly without effort. 
    }
    \label{fig:teaser}
    \vspace{-0.4cm}
\end{figure*}

It follows that rather than relying on writing (text) or drawing (sketch), all that is required is to choose the distinct visual concepts you specifically desire in your creative endeavor. Our model then ensures that all selected concepts are seamlessly and precisely composed into a faithful novel object in the final generation, without resorting to additional control such as bounding boxes \cite{xie2023boxdiff, bar2023multidiffusion, mo2023freecontrol, li2023gligen, zhang2023controlnet}. To illustrate with the ``unique bird'' example once more, our approach literally involves selecting the head of X, wings of Y, and body of Z! (see Fig.~\ref{fig:big-teaser}(e) and \ref{fig:teaser}).

Our solution is intuitive and centers around the well-studied computer vision concept of objects and their parts \cite{biederman1987recognition, felzenszwalb2005pictorial, krause2015fine, chen2019looks, he2023compositor}. The challenge then boils down to two aspects: (i) how to parse known objects into their parts (\ie, recognizing that birds have eyes and tails), and (ii) how to assemble parts from different visual concepts to form a faithful creative concept (\ie, ensuring the output is recognizably a bird).

The former is easier. We make clever use of DINOv2 \cite{oquab2023dinov2} feature maps and perform unsupervised clustering to identify common parts. The idea is that each cluster will then correspond to a semantic part of a common object (\eg, head and wings of birds). We specifically chose DINOv2 for its superior fine-grained perception \cite{amir2022effectiveness} compared to others \cite{peize2023vlpart,ren2024groundedsam, segrec2023groundedsam}. 
Further, this way has a higher flexibility to enhance fine-grained parsing by using a higher cluster count. 

Our major contribution lies in addressing the latter challenge. The solution is intuitive -- it essentially revolves around the fine-grained selection and placement of the chosen parts. Inspired by recent efforts in personalization \cite{ruiz2023dreambooth, gal2022textualinversion, wei2023elite, kumari2023multiconcept, avrahami2023breakascene}, primarily designed for learning entire objects, we introduce a tailored attention loss that specifically operates on parts. With this loss, our model learns the object part composition, ensuring the final generation appears holistically faithful with mixed parts (\ie, head and wings of a bird actually appear at the right places).

More specifically, we introduce an entropy-based attention loss that maximizes the attention of a specific part at a particular location while minimizing the attention where no parts appear. This is achieved by first selecting the attention maps that correspond to the parts. A normalization over all parts is then performed to ensure that each image region is occupied by no more than one part. Finally, we minimize the entropy loss between each normalized individual part and the semantic maps obtained during object parsing, containing the correct part location. This not only facilitates stronger part disentanglement but is also the key to generating a faithful holistic structure of an object as it learns a generic prior topology knowledge about object parts.

To enhance generation fidelity, we further employ a bottleneck encoder to project the text tokens. This approach accelerates learning by leveraging shared knowledge (common parts) and facilitating information exchange among instances in each part. Each instance adjusts slightly to adapt to the fine-grained part details during optimization.

Our contributions are as follows: 
\textbf{(i)} We pioneer a unique approach for fine-grained part-level control in Text-to-Image (T2I) models, empowering users to craft objects by selecting desired parts. This method streamlines the creative process, marking a significant advancement in our capacity to manipulate and reimagine visual content.
\textbf{(ii)} We introduce \textbf{\methodname{}}, a technique that autonomously parses object parts and orchestrates them from different visual concepts, resulting in the faithful creation of a novel object.
\textbf{(iii)} For enhanced part disentanglement and generation fidelity, we propose an entropy-based normalized attention loss and leverage a bottleneck encoder.
\textbf{(iv)} We present two problem-specific quantitative metrics. Comprehensive experiments on CUB-200-2011 (birds) and the Stanford Dogs dataset demonstrate the superior performance of our method in generating novel objects, surpassing alternative approaches in both qualitative and quantitative evaluations.


%% file: sec/2_related_work.tex
\vspace{-0.2cm}

\section{Related Work}

\noindent \textbf{Creative editing and generation.} Creativity involves generating innovative ideas or artifacts across various domains \cite{cetinic2022understanding}. Extensive research has explored the integration of creativity into Generative Adversarial Networks (GANs) \cite{ahmed2017can, nobari2021creativegan, sbai2019design} and Variational Autoencoders (VAEs) \cite{das2020creativedecoder, cintas2022creativity}. For example, DoodlerGAN \cite{ge2020doodlergan} learns and combines fine-level part components to create sketches of new species.
A recent study by \cite{vinker2023conceptdecomposition} demonstrated decomposing personalized concepts into distinct visual aspects, creatively recombined through diffusion models. InstructPix2Pix \cite{brooks2022instructpix2pix} allows creative image editing through instructions, while ConceptLab \cite{richardson2023conceptlab} aims to identify novel concepts within a specified category, deviating from existing concepts. 
Different from these works where editing/generation usually focuses on the whole object, we instead focus on training a text-to-image generative model that can understand parts, thus able to creatively generate new objects by seamlessly composing different parts simply through selection.

\vspace{0.1cm}

\noindent \textbf{Text-to-image generation.} Recent advancements in large text-to-image (T2I) diffusion models \cite{rombach2022ldm_sd, ramesh2022hierarchical_dalle, balaji2022ediffi, ding2022cogview2, nichol2021glide, saharia2022photorealistic_imagen, ding2021cogview, du2024demofusion} have made significant improvements over conventional methods \cite{zhu2019dmgan, tao2022dfgan, xu2018attngan, yin2019semantics, qiao2019mirrorgan, li2019controllable} in producing high-fidelity images from text prompts. Their application scope has expanded to include both global \cite{brooks2022instructpix2pix, mokady2023nulltext, kawar2023imagic} and localized \cite{couairon2023diffedit, yang2023paint, hertz2022prompttoprompt, avrahami2023blended} image editing tasks, demonstrating versatility.
Methods such as \cite{gafni2022makeascene, avrahami2023spatext, li2023gligen, zhang2023controlnet, bar2023multidiffusion, kim2023densediffusion, chefer2023attendandexcite, yang2023reco, feng2022structurediffusion, chen2024layoutguidance, xie2023boxdiff} have introduced more granular spatial control, such as incorporating semantic segmentation masks or bounding boxes, into large pretrained diffusion models to guide image generation. 
Contrary to these approaches, we focus on enhancing the control by enabling a straightforward discrete selection of desired parts. We minimize the complexity and manual intervention required by spatial controls, yet the model can compose selected parts as a coherent object seamlessly. 
This not only simplifies the user's role in the generative process but also ensures that the compositional logic and coherence are inherently managed by the model's ability. 
As such, our research is distinguished by focusing on the model's inherent ability to understand and apply part relationships instead of providing additional controls.

\vspace{-0.2cm}

\noindent \textbf{Abstracting visual knowledge as a text token.}
The effectiveness of T2I models is constrained by the user's ability to articulate their desired image through text. These models face challenges in faithfully replicating visual characteristics from a reference set and generating innovative interpretations in diverse contexts, even with detailed textual descriptions.
To address this challenge, various personalization techniques have been developed. These techniques obtain a new word embedding from multiple images depicting the same concept \cite{gal2022textualinversion, voynov2023pplus, alaluf2023neti, ruiz2023dreambooth, kumari2023multiconcept} or multiple new word embeddings for various concepts within a single image \cite{avrahami2023breakascene} through inversion. The learned visual concepts can be creatively reused in many image editing tasks. 
Nonetheless, most of these approaches struggle to learn object parts, often not able to follow the part selections due to part entanglement as they were designed to learn object as a whole.
In this work, we introduce a customized attention loss that serves a dual purpose: ensure accurate positioning of each part and enforce each image region occupied by no more than one part. This greatly improves the part disentanglement, creating novel concepts with correct appearances (see Fig.~\ref{fig:teaser}).

%% file: sec/3_methodology.tex
\section{Methodology}

\begin{figure}[t]
    \centering
    \includegraphics[width=\textwidth]{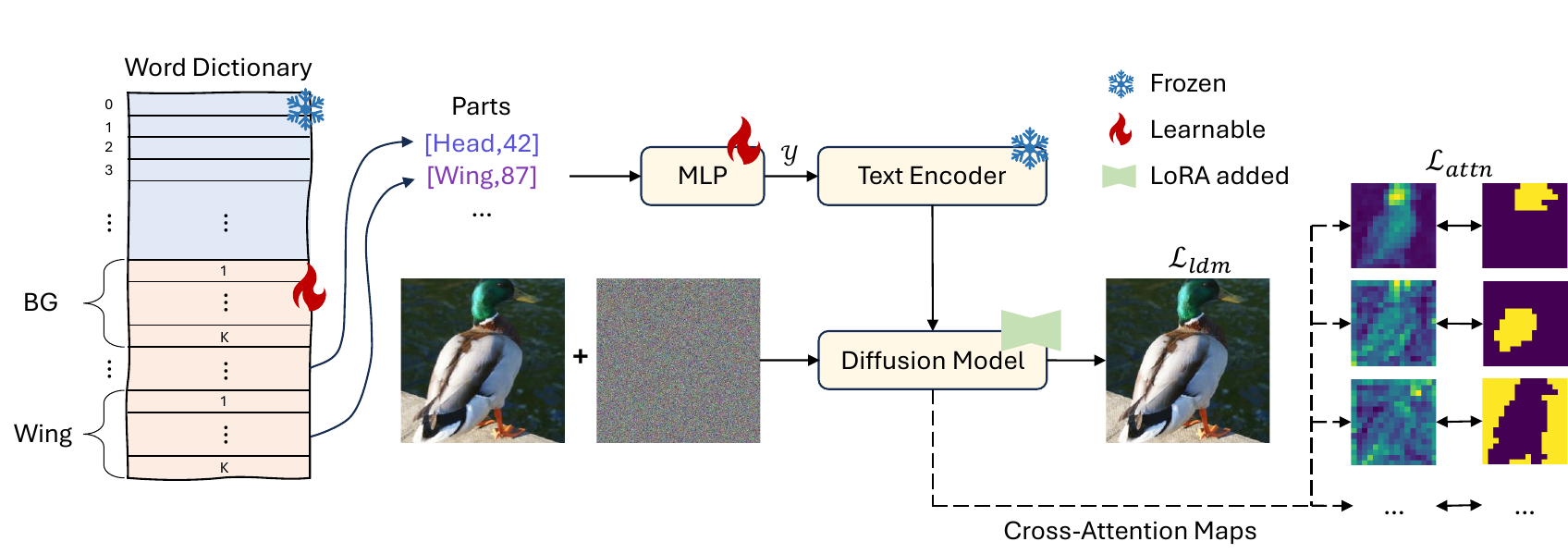}
    \caption{
    Overview of our \methodname{}. 
    All parts are organized into a dictionary, and their semantic embeddings are learned through a textual inversion approach. For instance, a text description like ``a photo of a [Head,42] [Wing,87]...'' guides the optimization of the corresponding textual embedding by reconstructing the associated image. To improve generation fidelity, we incorporate a bottleneck encoder $f$ (MLP) to compute the embedding $y$ (Eq.~\eqref{eq:token_embs}) as input to the text encoder. To promote disentanglement among learned parts, we minimize a specially designed attention loss, denoted as $\mathcal{L}_{attn}$.
    }
    \label{fig:coco-arch}
    \vspace{-0.4cm}
\end{figure}

Given a set of unlabeled images depicting the same object (\eg, bird) with different part details, we aim to train a T2I generative model that decomposes parts of objects into text tokens and can recompose them in a novel way. To that end, we propose \methodname{}, as depicted in Fig.~\ref{fig:coco-arch}.

We start by discovering the parts in a three-tier hierarchy, as detailed in Sec.~\ref{sec:kmeans}.
%
Paired with the training images $\{x_i\}^N_{i=1}$, this semantic hierarchy subsequently serves as the supervision to fine-tune a pre-trained text-to-image model, say a latent diffusion model \cite{rombach2022ldm_sd}, denoted as
$\{\epsilon_\theta, \tau_\theta, \mathcal{E}, \mathcal{D}\}$,
where $\epsilon_\theta$ represents the diffusion denoiser, $\tau_\theta$ the text encoder, and $\mathcal{E}/ \mathcal{D}$ the autoencoder respectively.
We adopt the textual inversion technique \cite{gal2022textualinversion}.
Concretely, we learn a set of pseudo-words $p^*$ for each part in the word embedding space with:
\begin{align} 
    \mathcal{L}_{ldm} &= \mathbb{E}_{z,t,p,\epsilon}\big[ || \epsilon - \epsilon_\theta(z_t,t,\tau_\theta(y_{p}))||^2_2 \big], \label{eq:diffusionloss} \\
    p^* &= \argmin_p \, \mathcal{L}_{ldm} \label{eq:textinv_obtain},
\end{align}
%
where $\epsilon \sim \mathcal{N}(0,1)$ denotes the unscaled noise, $t$ is the time step, $z = \mathcal{E}(x)$ is the latent representation of the image, $z_t$ is the latent noise at time $t$, and $y_{p}$ is the text condition that includes $p$ as part of the text tokens. $\mathcal{L}_{ldm}$ is a standard diffusion loss \cite{ho2020ddpm} to reconstruct the parts. As each object is composed of a set of parts, its reconstruction is achieved by the reconstruction of the associated set of parts. In other words, when all parts are reconstructed properly, it will become a valid object.

\vspace{-0.2cm}

\subsection{Unsupervised Part Discovery}
\label{sec:kmeans}








\begin{wrapfigure}{l}{0.45\textwidth}
    \centering
    \vspace{-0.65cm}
    \includegraphics[width=\linewidth]{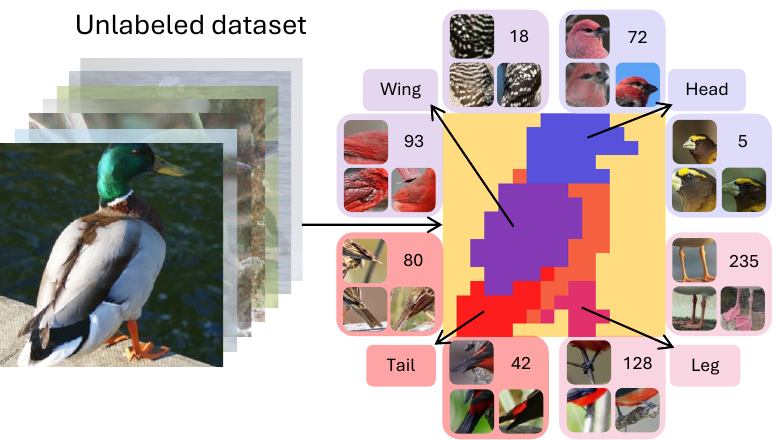}
    \caption{Part discovery within a semantic hierarchy involves partitioning each image into distinct parts and forming semantic clusters across unlabeled training data. } \label{fig:partdiscovery}
    \vspace{-0.55cm}
\end{wrapfigure}

To minimize the labeling cost, we develop a scalable process
to reveal the underlying semantic hierarchy with parts in an unsupervised fashion.
We leverage the off-the-shelf vision model for image decomposition and clustering.
Specifically, given an image $x_i$, we employ DINOv2 \cite{oquab2023dinov2} to extract the feature map $F=\{F_i = \mathrm{DINOv2}(x_i)\}_i^N$. 
We then conduct three-level hierarchical clustering (see Fig.~\ref{fig:partdiscovery}):
%
\textbf{(i)} At the top level, $k$-means is applied on all patches in $F$ with $k=2$ to separate foregrounds and backgrounds $B$. 
\textbf{(ii)} At the middle level, $k$-means is further applied on all foreground patches to acquire $M$ clusters representing common parts, such as the heads of birds. 
\textbf{(iii)} At the bottom level, we further group each of the $M$ clusters as well as the background cluster $B$ into $K$ splits. Each split refers to finer meanings, such as the head of a specific bird species, or a specific background style.
Lastly, each region of an image will be tagged with the corresponding cluster index.
We represent these cluster tags as follows:
\begin{align} \label{eq:concept}
    p = (0,k_0), (1, k_1), ..., (M, k_M),
\end{align}
where the first pair refers to the background style,
and the following $M$ pairs denote the combinations of 
$M$ parts (\eg, head, body, wings) each associated with a specific object (\eg, sparrow), and $k \in \{1, \ldots, K\}$.
This description will be used as the textual prompt in model training, such as ``a photo of a [$p$]''.
Please refer to the supplementary material for more examples of the discovered semantic hierarchy. This process also yields the segmentation mask of each $m$-th part, which we define as $S_m$. 

\noindent \textbf{Motivation.} While we can leverage off-the-shelf segmentation models such as VLPart \cite{peize2023vlpart}, the robustness relies on the generalizability of the model and the part segmentation result is usually pre-defined and may be unstable for unseen domains.
As a result, we rely on our feature clustering method to obtain the segmentation map, which also has a higher flexibility in choosing the number of clusters (parts).

\vspace{-0.2cm}

\subsection{Part Token Bottleneck}
\label{sec:mapper}

In contrast to prior text inversion studies \cite{gal2022textualinversion}, our task requires learning a greater quantity at the same time -- specifically, {$(M+1)K$}-of word tokens derived from a collection of discovered parts marked by inherent imperfections (such as partial overlap and over splitting). This makes the learning task more demanding. To enhance the learning process, we propose a neural network $f(\cdot)$ comprising a two-layer MLP with ReLU activation:
\begin{align} \label{eq:token_embs}
    y_p = f(e(p)),
\end{align}
where $y_p$ will be subsequently used as the input\footnote{Word templates such as ``a photo of a [*]'' will be used.} to the text encoder $\tau_\theta$ and
$e \in \mathbb{R}^{MK \times D}$ is a learnable word embedding dictionary that maps $p$ to their respective embeddings. 

Our design demonstrates quicker convergence than directly learning the final word embeddings $e(\cdot)$ \cite{gal2022textualinversion} (see Fig.~\ref{fig:training_iters}). This could be attributed to the entanglement of word embeddings in the conventional design, where there is no information exchange among them during optimization. For instance, each token doesn't know they are learning for a specific part of a specific species. This lack of communication leads to lower data efficiency and slower learning. With the bottleneck $f$, it will first project the token into a common part embedding space (\eg, head), then slightly adjust itself to adapt the fine-grained part details. It's worth noting that the conventional design is a specific instance of our approach when $f$ is an identity function.

\vspace{-0.2cm}

\subsection{Learning to Craft by Parts}\label{sec:learningobjective}




Fine-tuning the T2I model, rather than sorely learning pseudo-words, has been shown to achieve better reconstruction of target concepts as demonstrated in \cite{ruiz2023dreambooth, kumari2023multiconcept}. However, this comes with a significant training cost. Thus, we apply LoRA (low-rank adaptation) \cite{hu2022lora} to the cross-attention block for efficient training. We then minimize the diffusion loss $\mathcal{L}_{ldm}$ (Eq.~\eqref{eq:diffusionloss}) to learn both pseudo-words and LoRA adapters.

While training with only $\mathcal{L}_{ldm}$, entanglement happens between parts, as evident from the attention maps in the cross-attention block of the denoiser $\epsilon_\theta$ (see Fig.~\ref{fig:before_after_attn}). This entanglement arises due to the correlation between parts (\eg, a bird head code is consistently paired with a bird body code to represent the same species). To address this issue, we introduce an entropy-based attention loss as regularization:
\begin{align} \label{eq:attn_loss}
    \mathcal{L}_{attn} &= \mathbb{E}_{z,t,m} \big[ -\big(S_{m} \log \hat{A}_m + (1 - S_{m}) \log (1 - \hat{A}_m)\big) \big], \\
    \hat{A}_{m,i,j} &= \frac{\bar{A}_{m,i,j}}{\sum_k \bar{A}_{k,i,j}}, \quad  \bar{A}_m = \frac{1}{L}\sum_l^L A_{l,m}, 
\end{align}
where $A \in [0,1]^{M \times HW}$ represents the cross-attention map between the $m$-th part and the noisy latent $z_t$, $L$ represents the number of specific layers to select attention maps, $\hat{A} \in [0,1]^{M \times HW}$ represents the averaged and normalized cross-attention map over all parts and $S_m \in \{0,1\}^{M \times HW}$ serves as the mask that indicates the location of $m$-th part. In cases where the part is not present in the image (\eg, occluded), we set both $S_m$ and $\hat{A}_m$ as 0 to exclude them.
Thus, the overall learning objective is defined as:
\begin{align}
    \mathcal{L}_{total} = \mathcal{L}_{ldm} + \lambda_{attn} \mathcal{L}_{attn},
\end{align}
where $\lambda_{attn}=0.01$. 
We focus on attention maps at the resolution of $16 \times 16$ where rich semantic information is captured \cite{hertz2022prompttoprompt}. Normalization is performed at each location to ensure that the sum of a patch location equals 1. This aims to maximize the attention of a specific part at a particular location which implicitly minimizing the attention of other parts similar to a softmax classification task. Compared to the mean-square based attention loss \cite{avrahami2023breakascene}, this intuitively ensures that a part only appears once at a particular location, facilitating stronger disentanglement from other parts during the denoising operation. When generating a part for a particular location, the diffusion model $\epsilon_\theta$ should only attend to the part instead of other non-related parts.

%% file: sec/4_experiment.tex
\vspace{-0.2cm}

\section{Experiments}

\noindent \textbf{Datasets.} We demonstrate our  selection task on two fine-grained object datasets: CUB-200-2011 (birds) \cite{wah2011cub200} which contains 5,994 training images, and Stanford Dogs \cite{khosla2011dog} which contains 12,000 training images.



\vspace{0.1cm}

\noindent \textbf{Implementation.} 
For part composition, we assess the model's ability to combine up to 4 different parts from 4 distinct species/objects. 
We set $M=5$ for bird generation (head, front body/breast area, wings, legs, tail) and $M=7$ for dog generation (forehead, eyes, mouth/nose, ears, neck, body/tail, legs). For both datasets, $K$ is set as 256, ensuring sufficient coverage of all fine-grained classes (\ie, 200 for birds and 120 for dogs). 
We randomly generate 500 images by sampling 500 sets of parts. For each image, we randomly replace an original part with any part from another 500 non-overlapping sets of parts. The resulting set of parts may take the form of ``$(0,k_A)$ $(1,k_B)$ $(2,k_C)$ ... $(M,k_D)$'', representing a composition from species A, B, C, and D. 
Stable Diffusion v1.5 \cite{rombach2022ldm_sd} is used. Please see the supplementary material for further training details.



\begin{table}[t]
    \centering
    \adjustbox{max width=0.9\linewidth}{
        \begin{tabular}{c|cccc}
            Method & ~Learnable Token~ & ~Fine Tune~ & ~Disentanglement~ & ~Bottleneck~ \\
             \midrule
            Textual Inversion \cite{gal2022textualinversion} & \cmark & \xmark & \xmark & \xmark \\
            DreamBooth \cite{ruiz2023dreambooth}  & \cmark & LoRA* & \xmark & \xmark \\
            CustomDiffusion \cite{kumari2023multiconcept}  & \cmark & $K/V$ & \xmark & \xmark \\
            Break-a-scene \cite{avrahami2023breakascene}  & \cmark & LoRA*  & MSE & \xmark \\
            PartCraft (Ours) & \cmark & LoRA & Eq.~\eqref{eq:attn_loss} & \cmark \\
        \end{tabular}
    }
    \vspace{0.1cm}
    \caption{Comparing our and alternative methods in design properties.
    *: We fine-tuned the added LoRA \cite{hu2022lora} adapter rather than the entire diffusion model $\epsilon_\theta$ due to resource limit. MSE is a mean-square based attention loss used in \cite{avrahami2023breakascene}.
    }
    \vspace{-0.5cm}
    \label{tab:summary_methods}
\end{table}

We compare our method with the recent personalization methods: Textual Inversion (TI) \cite{gal2022textualinversion}, DreamBooth (DB) \cite{ruiz2023dreambooth}, Custom Diffusion (CD) \cite{kumari2023multiconcept}, Break-a-scene (BaS) \cite{avrahami2023breakascene}. 
These personalization methods were designed to take single or multiple images with associated labeled objects as input. 
The text prompt for each image is as simple as ``a photo of [$p$]'' where $p$ is expressed in Eq.~\eqref{eq:concept}, since we do not rely on complex prompts.
We employ the official implementations released by the original authors for training.
We summarize the main design properties of all compared methods in Tab.~\ref{tab:summary_methods}.

\vspace{0.1cm}

\noindent \textbf{Evaluation metrics.} To assess a model's ability to disentangle and composite parts, we introduce two metrics: (a) exact matching rate (\textit{EMR}) and (b) cosine similarity (\textit{CoSim}) between the $k$-means embeddings of the parts of real and generated images. Utilizing the pre-trained $k$-means from Sec.~\ref{sec:kmeans}, we predict the parts of generated images. \textit{EMR} quantifies how accurately the cluster index of parts of generated images matches the parts of the corresponding real images whereas \textit{CoSim} measures the cosine similarity between the $k$-means centroid vector that the part belongs to between generated and real images.
These metrics assess the model's ability to follow the input parts and accurately reconstruct them, with perfect disentanglement indicated by \textit{EMR} of 1 and \textit{CoSim} of 1. A detailed algorithm is provided in the supplementary material.
We also measure image generation quality using FID \cite{heusel2017gans_fid} to assess model performance in terms of image distribution. 
Additionally, we compute the average pairwise cosine similarity between CLIP \cite{radford2021clip}/DINO \cite{caron2021emergingdino} embeddings of generated and real class-specific images following \cite{ruiz2023dreambooth}. 
Each generated image is conditioned on the parts of the corresponding real image. This results in 5,994 generated images for birds and 12,000 generated images for dogs.

\vspace{-0.4cm}

\subsection{Part Reconstruction and Image Quality Evaluation} \label{sec:conveval}

We first assess the ability of different methods to learn parts as text tokens by evaluating how well they can accurately reconstruct the parts (this also means image generation with original parts).

\begin{table}[t]
    \centering
    \adjustbox{max width=\linewidth}{
        \begin{tabular}{c|ccccc|ccccc}
            \multirow{2}{*}{Method}& \multicolumn{5}{c|}{Birds: CUB-200-2011} & \multicolumn{5}{c}{Dogs: Stanford Dogs} \\
            \cmidrule{2-11}
             & FID & CLIP & DINO & EMR & CoSim & FID & CLIP & DINO & EMR & CoSim \\
            \midrule
            Textual Inversion \cite{gal2022textualinversion}  & \bf 10.10 & \bf 0.784 & 0.607 & 0.305 & 0.842 & 23.36 & 0.652 & 0.532 & 0.218 & 0.754 \\
            DreamBooth  \cite{ruiz2023dreambooth} & 12.94 & 0.775 & 0.594 & 0.355 & 0.856 & 22.65 & 0.660 & 0.563 & 0.275 & 0.777 \\
            Custom Diffusion \cite{kumari2023multiconcept} & 37.61 & 0.694 & 0.504 & 0.338 & 0.833 & 42.41 & 0.593 & 0.491 & 0.253 & 0.755 \\
            Break-a-Scene \cite{avrahami2023breakascene} & 20.05 & 0.742 & 0.549 & 0.390 & 0.854 & 24.20 & 0.633 & 0.532 & 0.300 & 0.775 \\
            \midrule
            \methodname{} (Ours) & 12.86 & 0.783 & \bf 0.618 & \bf 0.460 & \bf 0.882 & \bf 16.92 & \bf 0.669 & \bf 0.573 & \bf 0.358 & \bf 0.796 \\
        \end{tabular}
    }
    \vspace{0.1cm}
    \caption{Quantitative comparison for part reconstruction.}
    \vspace{-0.5cm}
    \label{tab:birddog_score}
\end{table}

\begin{figure}[t]
    \centering    
    \begin{subfigure}[b]{0.192\linewidth}
        \includegraphics[width=\linewidth]{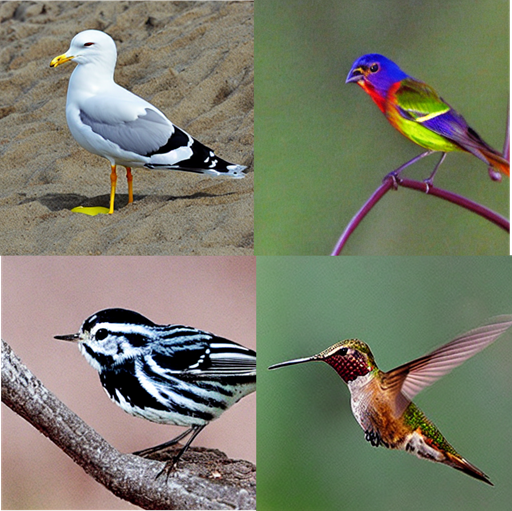} \\
        \includegraphics[width=\linewidth]{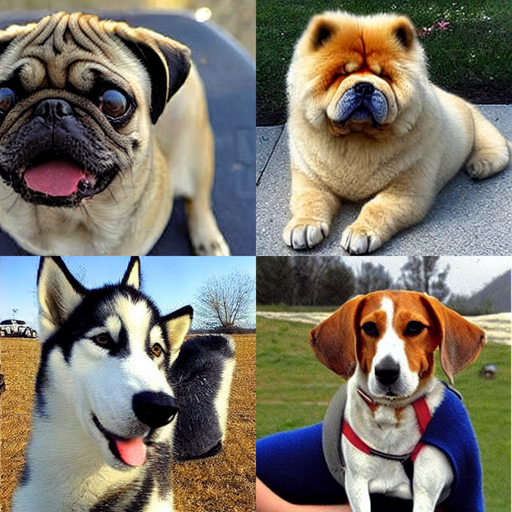}
        \caption{TI}
    \end{subfigure}
    \begin{subfigure}[b]{0.192\linewidth}
        \includegraphics[width=\linewidth]{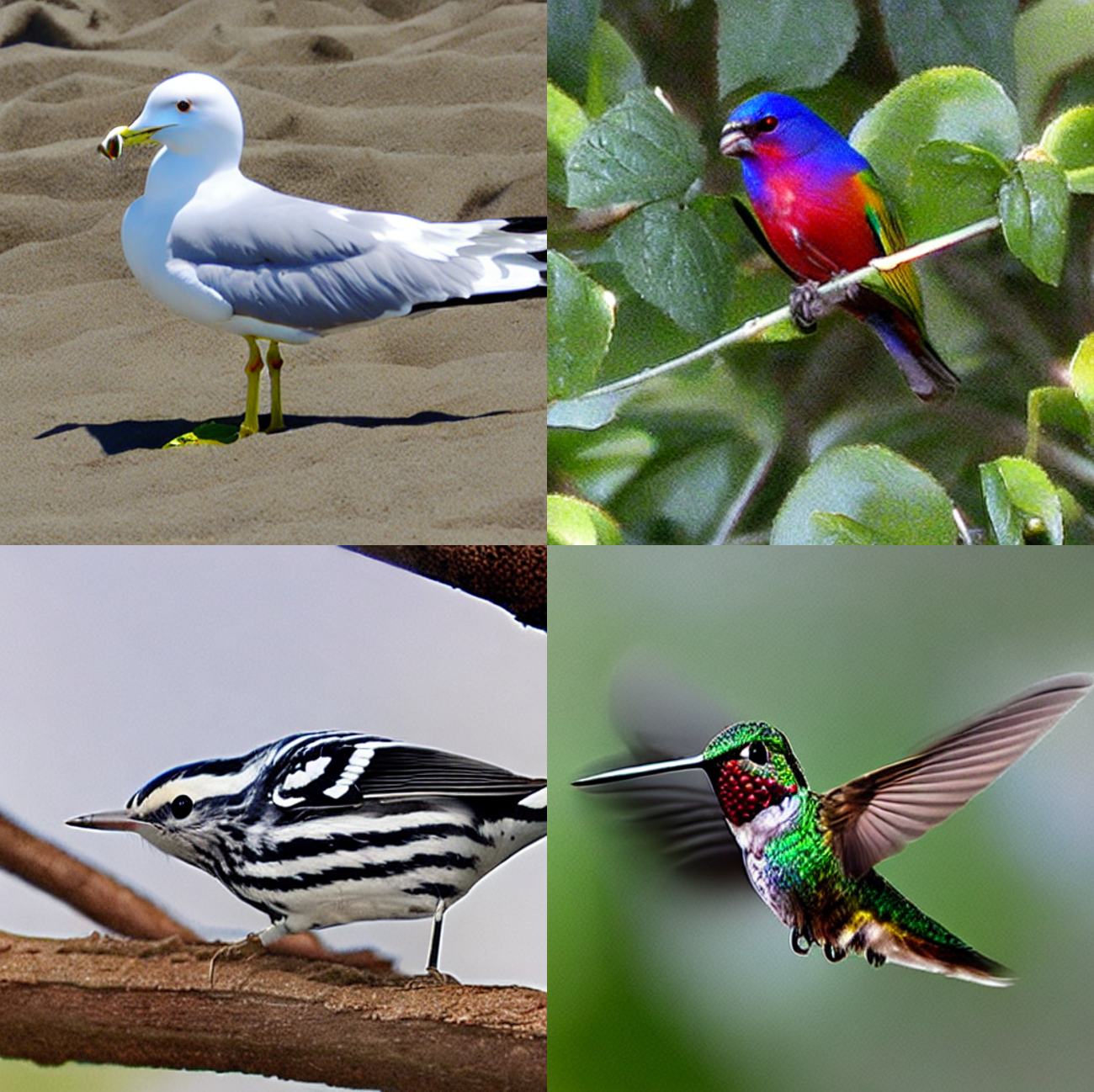} \\   
        \includegraphics[width=\linewidth]{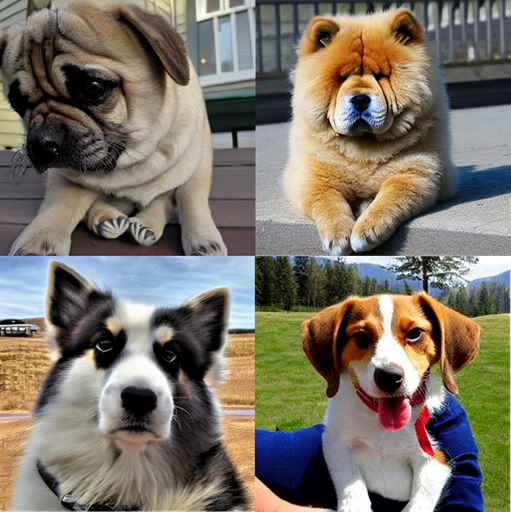}
        \caption{DB}
    \end{subfigure}
    \begin{subfigure}[b]{0.192\linewidth}
        \includegraphics[width=\linewidth]{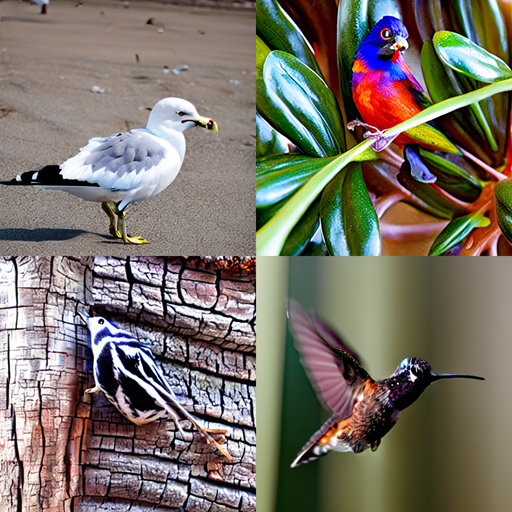} \\   
        \includegraphics[width=\linewidth]{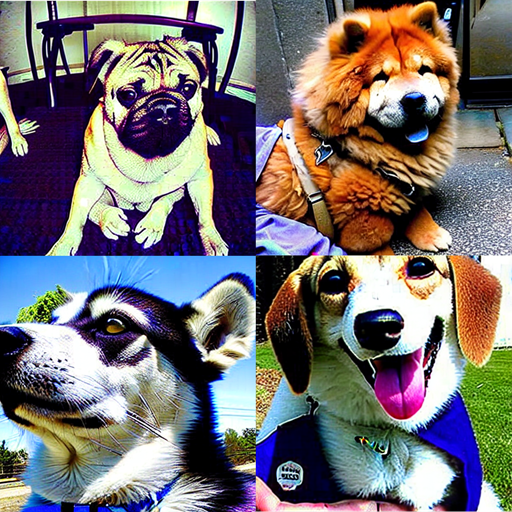}
        \caption{CD}
    \end{subfigure}
    \begin{subfigure}[b]{0.192\linewidth}
        \includegraphics[width=\linewidth]{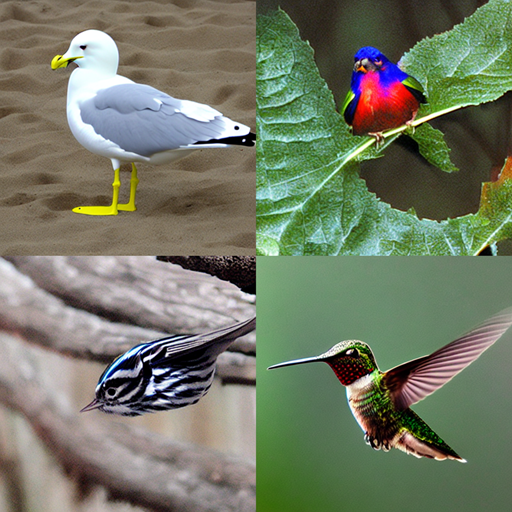} \\   
        \includegraphics[width=\linewidth]{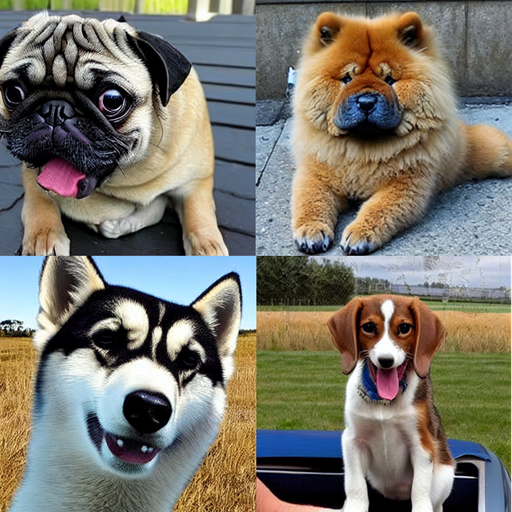}
        \caption{BaS}
    \end{subfigure}
    \begin{subfigure}[b]{0.192\linewidth}
        \includegraphics[width=\linewidth]{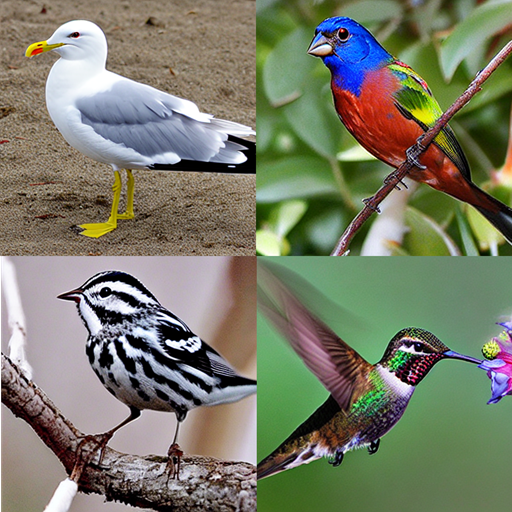} \\ 
        \includegraphics[width=\linewidth]{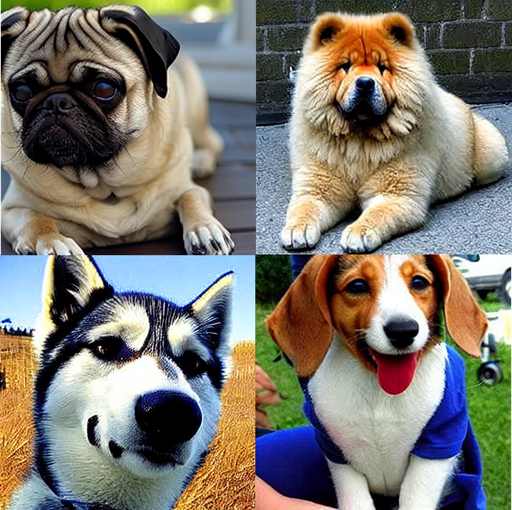}
        \caption{\methodname{}}
    \end{subfigure}
    \vspace{-0.1cm}
    \caption{Visual comparison under the part reconstruction setting. All images are generated by using the original parts of respective objects. 
    }
    \vspace{-0.2cm}
    \label{fig:class_specific_generated_image}
\end{figure}

In Tab.~\ref{tab:birddog_score}, we summarize the performance of respective methods on the bird and dog generation, respectively. We highlight four observations:
\textbf{(i)} Textual Inversion performs quite well compared to DreamBooth, CustomDiffusion, and Break-a-scene in terms of FID, CLIP, and DINO scores although did not fine-tune the diffusion model $\epsilon_\theta$. This may be due to the potential risk of overfitting when fine-tuning $\epsilon_\theta$ especially when learning a vast array of new concepts with many update iterations. 
It is also not uncommon to carefully tune the learning rate and the training iterations in these models when fine-tuning new concepts (\eg, only 800-1000 steps of updates to learn a new concept in \cite{avrahami2023breakascene}).
\textbf{(ii)} Nonetheless, fine-tuning the diffusion model $\epsilon_\theta$ can help improve the ability to follow prompts as shown by increased EMR and CoSim scores (\eg, EMR of at least 5\% in DreamBooth).
\textbf{(iii)} Break-a-scene has a better ability to reconstruct the parts as shown by EMR and CoSim, this is due to the attention loss explicitly forcing the parts to focus on the respective semantic region.
\textbf{(iv)} 
\methodname{} achieves the best performance in DINO, EMR, and CoSim scores (\eg, 7\% better in EMR compared to Break-a-scene). This indicates that not only does our image-generation ability perform comparably well with textual inversion, but \methodname{} is also able to disentangle the parts learning so that it can follow the prompt instructions more accurately to generate the parts in a cohort. 
In Fig.~\ref{fig:class_specific_generated_image}, we present generated images from different methods, with CustomDiffusion exhibiting high-contrast images, possibly due to unconstrained fine-tuning on the cross-attention components $K/V$ and resulting in worse FID scores. 

\begin{figure}[t!]
    \centering
    \includegraphics[width=\linewidth]{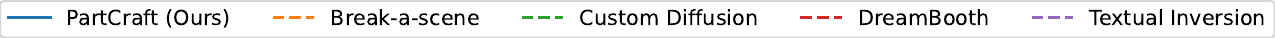} \\
    \begin{subfigure}[b]{0.49\textwidth}
         \centering
         \includegraphics[width=0.49\linewidth]{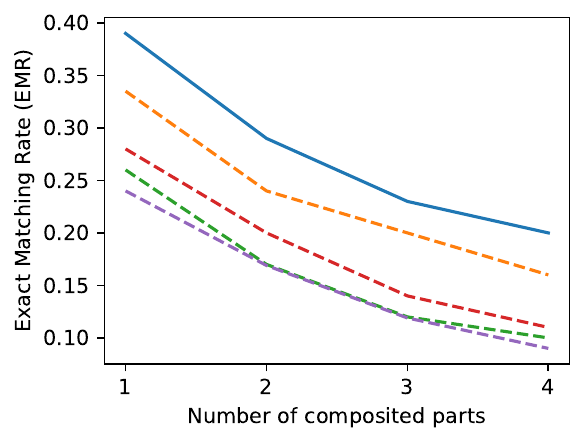}
         \hfill
         \includegraphics[width=0.49\linewidth]{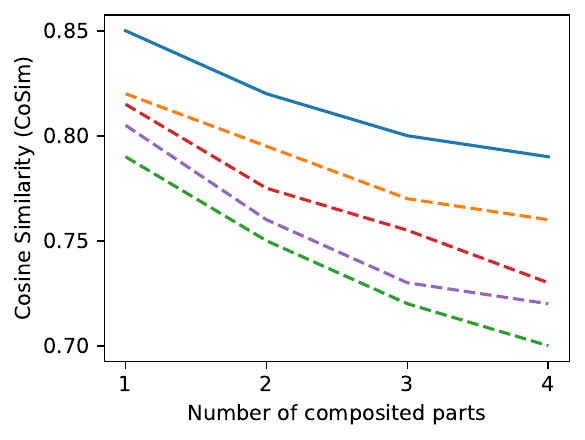}
         \caption{CUB-200-2011}
         \label{fig:bird_composition}
     \end{subfigure}
     \hfill
     \begin{subfigure}[b]{0.49\textwidth}
         \centering
         \includegraphics[width=0.49\linewidth]{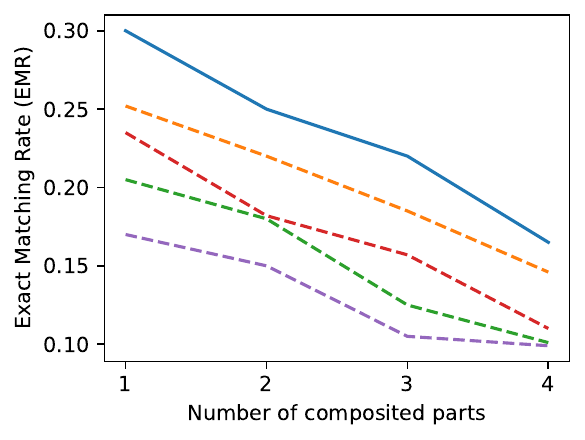}
         \hfill
         \includegraphics[width=0.49\linewidth]{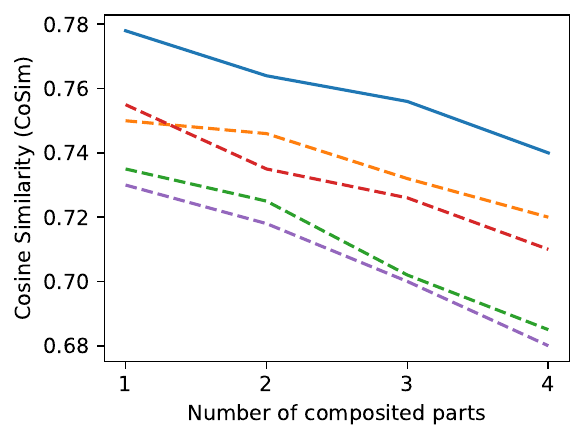}
         \caption{Stanford Dogs}
         \label{fig:dog_composition}
     \end{subfigure}
    \caption{Quantitative comparisons of part composition in terms of EMR and CoSim.}
    \vspace{-0.5cm}
    \label{fig:composition_compare}
\end{figure}


\begin{figure}[t!]
    \centering
    \includegraphics[width=\textwidth, height=0.7\textheight]{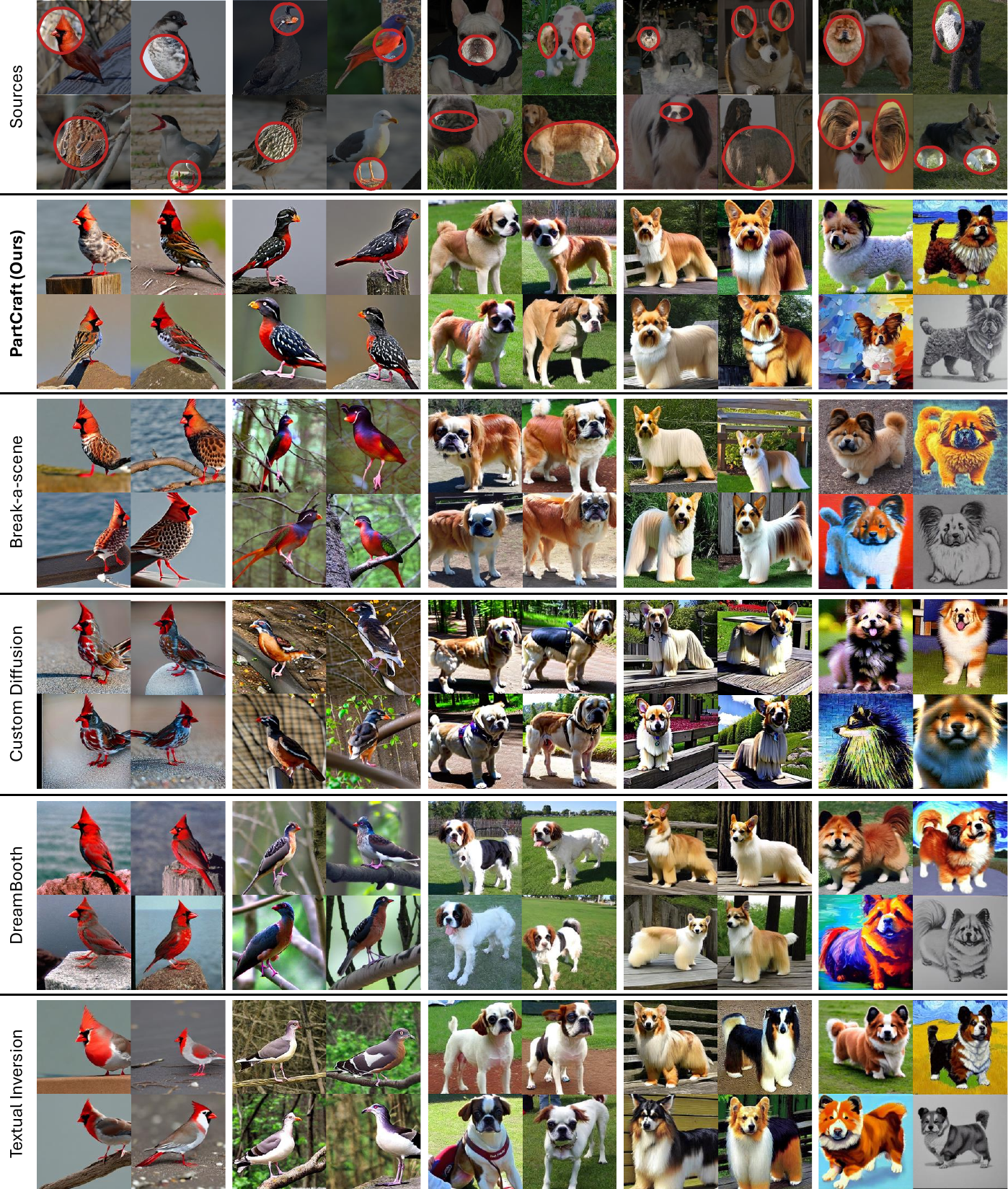}
    \captionof{figure}{Visual comparison on 4-species (specified on the top row) mixed generation. The last column indicates generated images with different styles (\ie, \textit{DSLR}, \textit{Van Gogh}, \textit{Oil Painting}, \textit{Pencil Drawing}). }
    \vspace{-0.6cm}
    \label{fig:composition_visualization}
\end{figure}

\vspace{-0.2cm}

\subsection{Part Composition Evaluation} \label{sec:vcceval}

In this section, we assess the part composition ability of different methods. In this experiment, we generate the image by mixing different parts from different species.
%
Our findings, as shown in Fig.~\ref{fig:composition_compare}, can be summarized as follows:
\textbf{(i)} As the number of composited parts increases, EMR and CoSim decrease, reflecting the challenge of composing multiple diverse parts.
\textbf{(ii)} Break-a-scene and \methodname{} achieve notably higher EMR and CoSim scores, thanks to disentanglement through attention loss minimization.
\textbf{(iii)} \methodname{} outperforms Break-a-scene significantly by token bottleneck and tailored attention loss.


In Fig.~\ref{fig:composition_visualization}, we visualize the results of composing 4 different parts. While all images appear realistic, most methods struggle to assemble all 4 parts. For instance, Break-a-scene missed out on the flurry body of \textit{kerry blue terrier} (rightmost column). In contrast, our methods successfully combine 4 different parts from 4 different species, demonstrating the superior ability of our approach to part composition. 
We also visualize additional examples of our method in the supplementary material. 

Furthermore, we explore the versatility of the adapted model by generating images with simple styles such as \textit{pencil drawing}. While most methods successfully incorporate specific styles into the generated image, Custom Diffusion often fails to do so, possibly due to the unconstrained fine-tuning of the cross-attention components $K/V$.



\begin{figure}[t] %
    \centering
    \vspace{0.1cm}
    \begin{minipage}[b]{0.49\textwidth}
        \begin{minipage}{\linewidth}
            \centering
            \includegraphics[width=\linewidth]{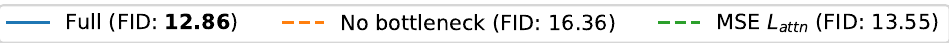} 
            \includegraphics[width=0.49\linewidth]{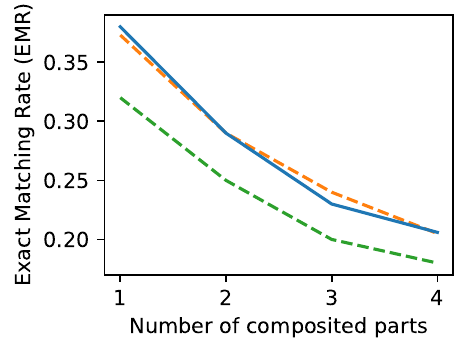}
             \hfill
             \includegraphics[width=0.49\linewidth]{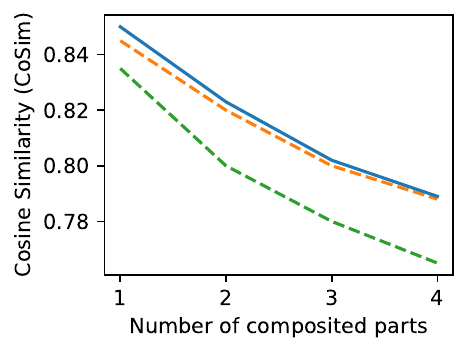}
             \captionof{figure}{Ablation on our part token bottleneck and attention loss
         under the part composition on CUB-200-2011 birds. } \label{fig:abl_comparison}
        \end{minipage}
        \begin{minipage}{\linewidth}
            \centering
            \vspace{0.2cm}
            \includegraphics[width=\linewidth]{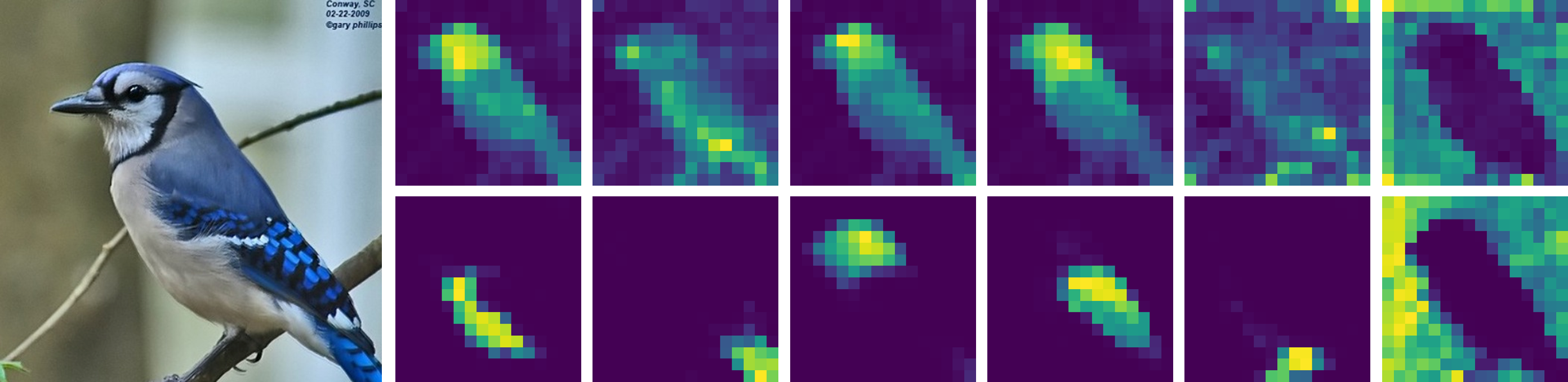}
            \captionof{figure}{Cross-attention map of each part (top) without and (bottom) with our attention loss.} \label{fig:before_after_attn}
        \end{minipage}
    \end{minipage}
    \hfill
    \begin{minipage}[b]{0.49\textwidth}
        \includegraphics[width=\linewidth]{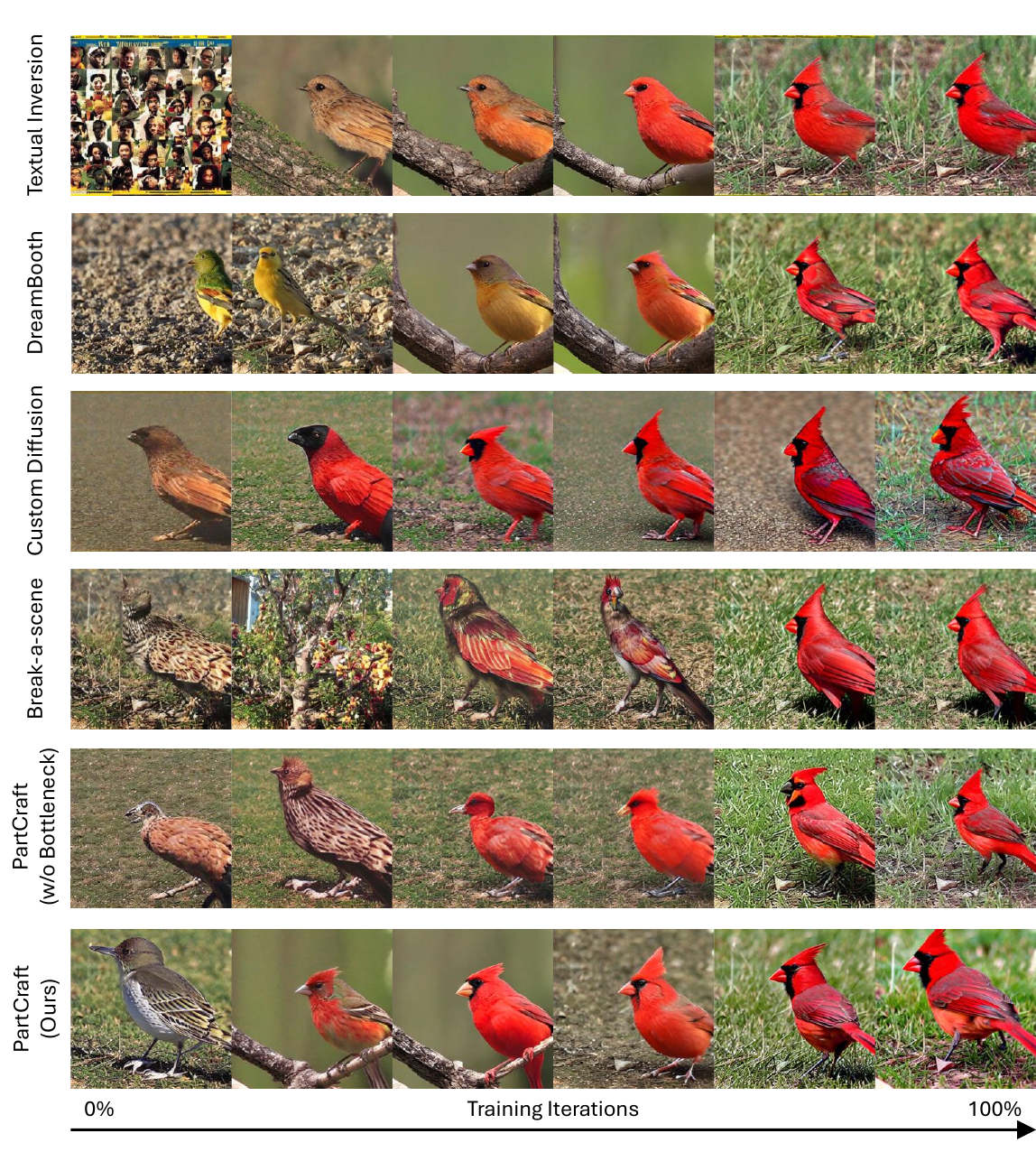} 
        \captionof{figure}{Generated images of \textit{cardinal} over different stages of training.}
        \vspace{-1.5cm}
        \label{fig:training_iters}
    \end{minipage}
\end{figure}

\subsection{Ablation Studies} \label{sec:abl}

\noindent \textbf{Component analysis.} 
In Fig.~\ref{fig:abl_comparison}, 
we evaluate the effect of our proposed components (token bottleneck and attention loss) on creating novel bird species. \textbf{(i)} Removing the bottleneck outlined in Eq.~\eqref{eq:token_embs} degrades the generation quality as evidenced by a higher FID score (12.86 $\rightarrow$ 16.36) even though both EMR and CoSim remain.
\textbf{(ii)} By replacing our $\mathcal{L}_{attn}$ with the MSE loss as proposed in \cite{avrahami2023breakascene}, we observe significant deterioration in both EMR and CoSim. \textbf{(iii)} Finally, incorporating both the projector and our attention loss performs the best. This improvement highlights the necessity of incorporating interactions between multiple parts to achieve more effective part disentanglement and optimization. 


\noindent \textbf{Cross-attention visualization.} Our attention loss plays a crucial role in token disentanglement.
We demonstrate the impact of this loss in 
Fig.~\ref{fig:before_after_attn},
where we observe significantly enhanced disentanglement after explicitly guiding attention to focus on distinct semantic regions. 

\begin{figure}[t]
    \centering
    \begin{subfigure}[b]{0.32\linewidth}
        \includegraphics[width=\linewidth]{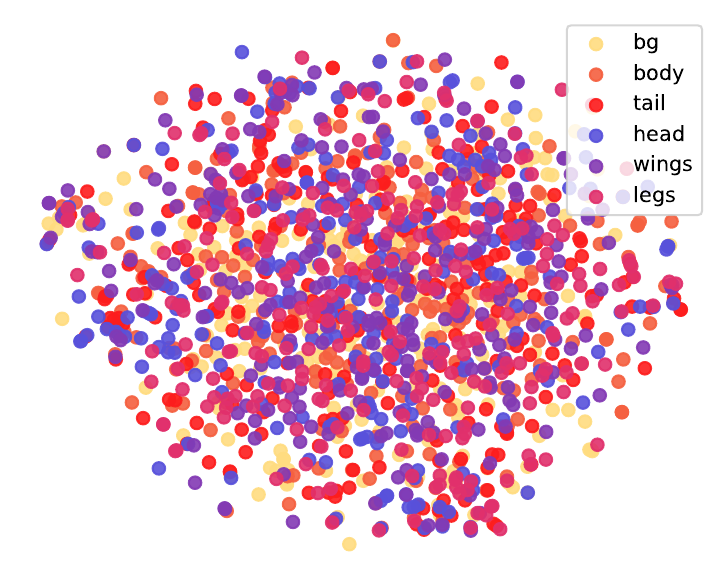}
        \caption{Textual Inversion}
    \end{subfigure}
    \hfill
    \begin{subfigure}[b]{0.32\linewidth}
        \includegraphics[width=\linewidth]{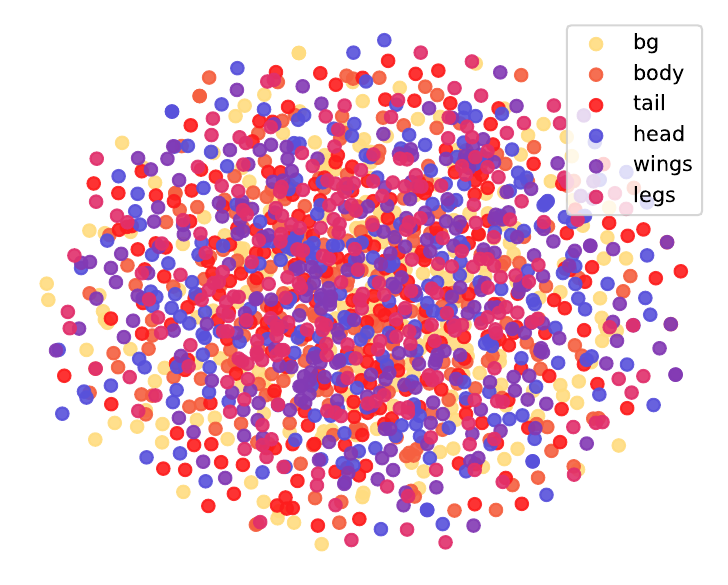}
        \caption{Break-a-scene}
    \end{subfigure}
    \hfill
    \begin{subfigure}[b]{0.32\linewidth}
        \includegraphics[width=\linewidth]{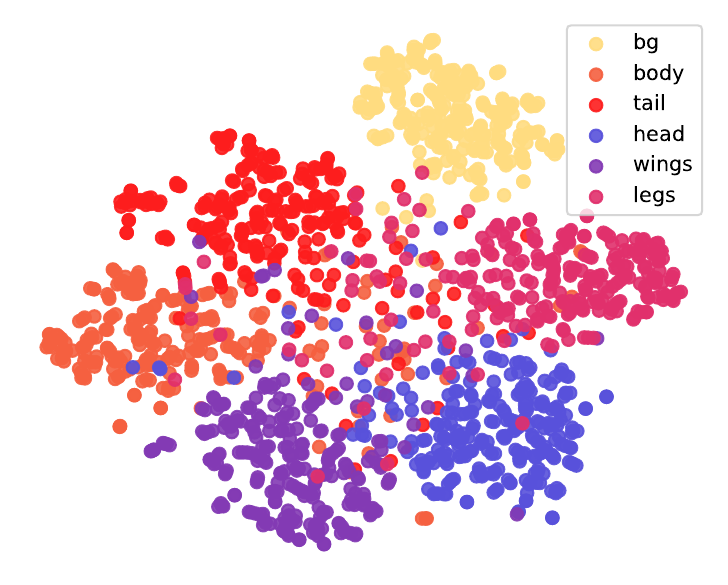}
        \caption{\methodname{} (Ours)}
    \end{subfigure}
    \caption{2D tSNE \cite{van2008tsne} projection of word embeddings. Different colors represent different common parts. (correspond to the segmentation mask in Fig.~\ref{fig:partdiscovery}).}
    \label{fig:tsne}
    \vspace{-0.5cm}
\end{figure}


\noindent \textbf{Part word embedding space.} We visualize the word embeddings of learned tokens of Textual Inversion \cite{gal2022textualinversion}, Break-a-scene \cite{avrahami2023breakascene} and our \methodname{} for birds generation (CUB-200-2011 \cite{wah2011cub200}) through tSNE \cite{van2008tsne} in Fig.~\ref{fig:tsne}. In our \methodname{}, the word embeddings are the projected embeddings through Eq.~\eqref{eq:token_embs}. 
We can see that our projected version has a better semantic meaning such that the part embeddings are clustered together by their semantic meaning (\eg, head). We believe this is one of the reasons that our \methodname{} outperforms previous methods in which we can compose all parts seamlessly yet with higher quality.


\noindent \textbf{Convergence analysis.} We present a visual comparison of images generated by various methods in Fig.~\ref{fig:training_iters}, spanning from the initial to the final stages of training. Notably, our \methodname{} demonstrates an ability to learn new concepts at even the early stages of training. Without the bottleneck encoder, we observe that the learning speed drops significantly (as evidenced by generating wrong part detail), indicating how the bottleneck serves as an important component when learning new tokens that have shared properties (\eg, common part).

\begin{figure}[t]
    \centering
        \includegraphics[width=0.18\linewidth]{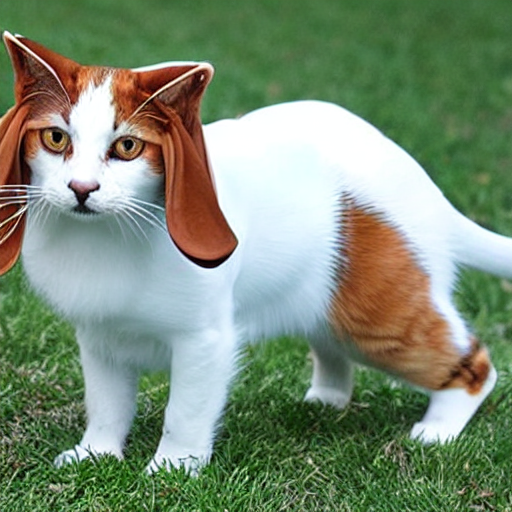}
        \includegraphics[width=0.18\linewidth]{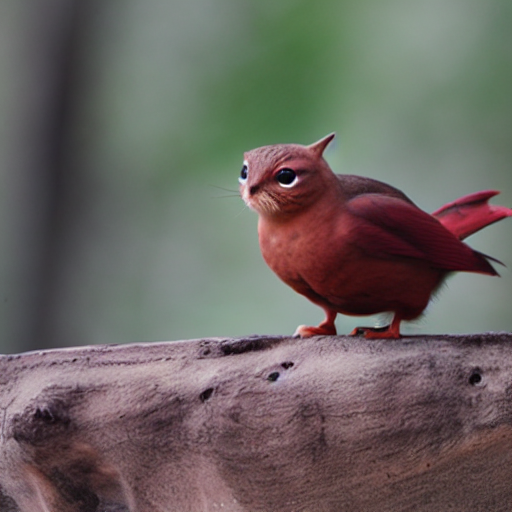}
        \includegraphics[width=0.18\linewidth]{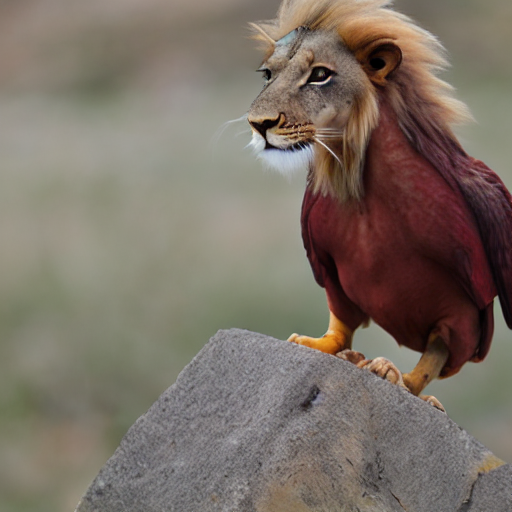}
        \includegraphics[width=0.18\linewidth]{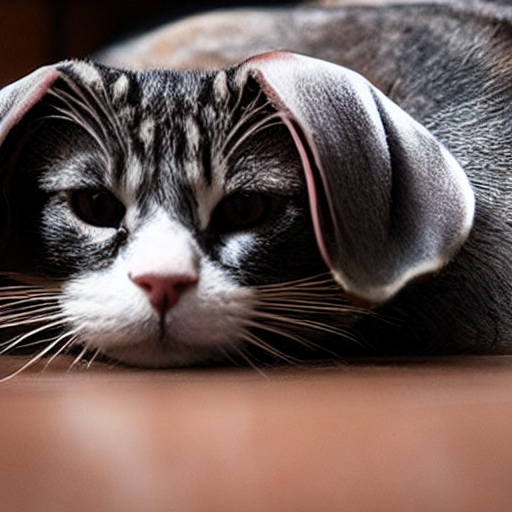}
        \includegraphics[width=0.18\linewidth]{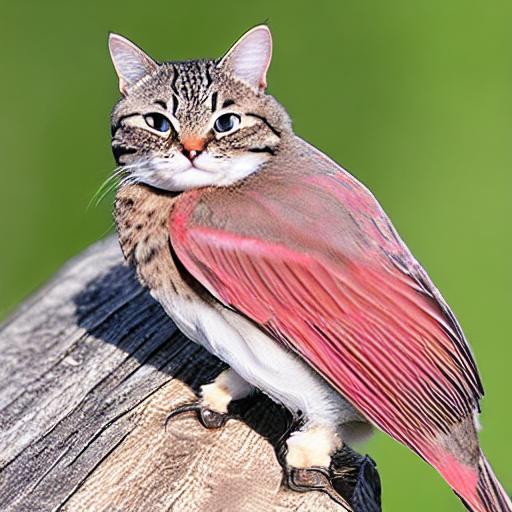}
        \includegraphics[width=0.18\linewidth]{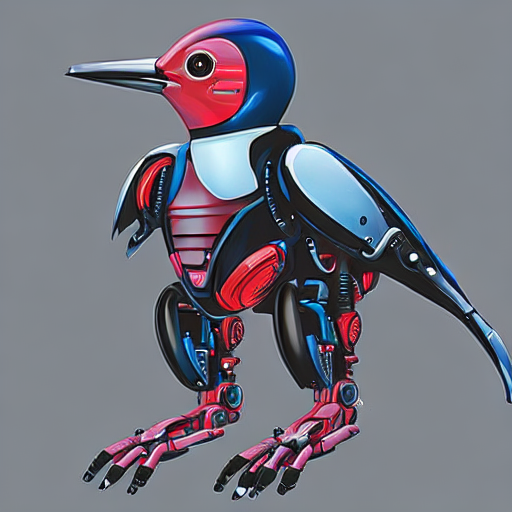}
        \includegraphics[width=0.18\linewidth]{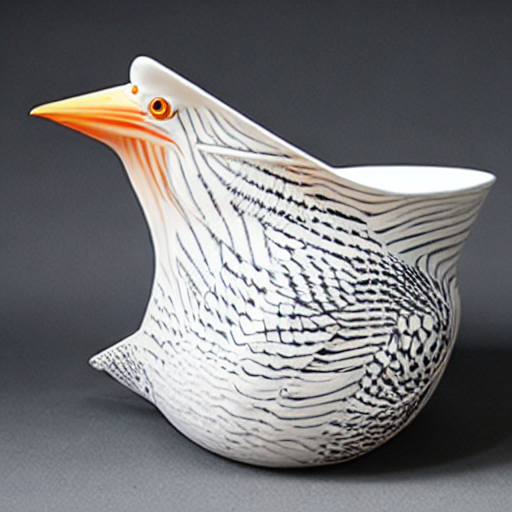}
        \includegraphics[width=0.18\linewidth]{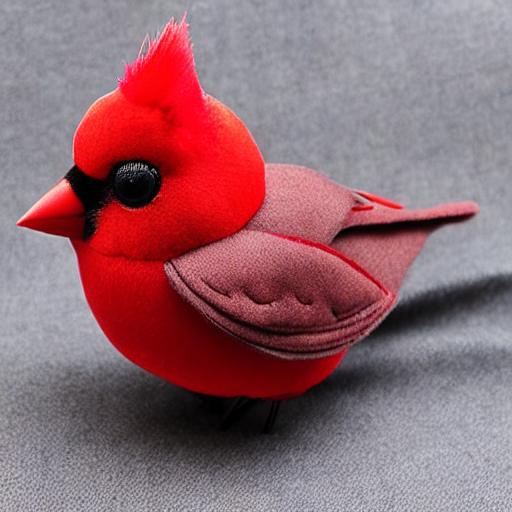}
        \includegraphics[width=0.18\linewidth]{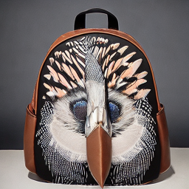}
        \includegraphics[width=0.18\linewidth]{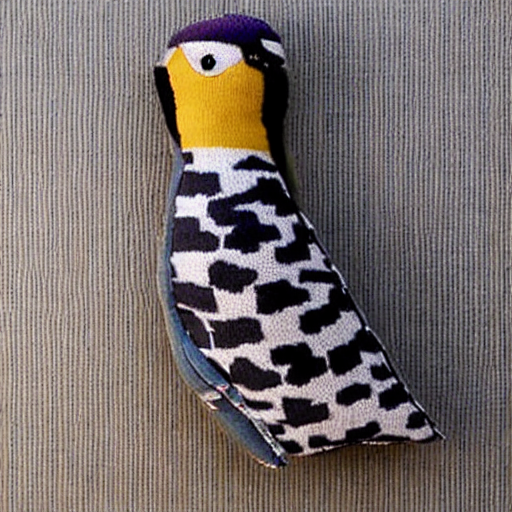}
    \caption{\textbf{(Top)}: Using learned parts to modify the property of other domains such as cat, and lion with prompt such as ``A cat with [\textit{beagle's ear}]''. \textbf{(Bottom)} We can also repurpose learned parts for creative image generation using prior knowledge in Stable Diffusion with prompt such as ``A robot designed inspired by [\textit{red header woodpecker's head}] and [\textit{blue jay's body}]''.}
    \label{fig:application}
    \vspace{-0.4cm}
\end{figure}


\noindent \textbf{Transferability for Creativity.} \textbf{(i)} In Fig.~\ref{fig:application}, we demonstrate that not only it can compose parts within the training domain (\eg, birds), but it can also transfer the learned parts to and combine with other domains (\eg, cat). This enables the creation of unique combinations, such as a cat with a dog’s ear. \textbf{(ii)} Leveraging the prior knowledge embedded in Stable Diffusion, \methodname{} can also repurpose learned parts for creative image generation. An example of this is the generation of a bird-shaped robot adorned with various parts. These examples showcase \methodname{}'s immense potential for diverse and limitless creative applications. Please see the supplementary material for more examples.

%% file: sec/5_conclusion.tex
\section{Conclusion}

\vspace{-0.1cm}

We propose a new way of control in generative AI. Instead of text or sketch, we ``select'' desired parts to create an object. We addressed the challenge of learning parts in T2I models by introducing a customized attention loss. This loss serves a dual purpose: to ensure parts are at the right location and to ensure each location is occupied by not more than one part. This greatly improves the part disentanglement. We further employ a non-linear bottleneck encoder to improve generation fidelity.
Our model, \methodname{}, can seamlessly compose different parts from different objects, creating objects that do not exist yet holistically correct and plausible objects by mixing them. 
Extensive experiments demonstrated \methodname{}'s superior performance in both qualitative and quantitative evaluation. Moreover, the learned parts demonstrate strong transferability. 
We hope that our \methodname{} will empower artists, designers, and enthusiasts to bring the creations of their dreams to reality.

\vspace{-0.2cm}

\section{Limitations and Future Works}

It is worth noting that the accuracy of obtained parts may be affected by using a self-supervised pre-trained feature extractor. Future work may explore the incorporation of encoders, such as \cite{wei2023elite}, to improve part accuracy. We also observed challenges in composing relatively small parts, like tails and legs, which require further investigation. Additionally, we are also exploring cross-domain generation, \ie, combining learned parts from different datasets to create objects with even more diverse parts. For instance, we can merge non-rigid parts (\eg, parts from quadrupled animals) with rigid parts (\eg, parts from cars/airplanes) and form a creative structure (\eg, a car that has horse legs instead of wheels). This not only further improves the applicability of \methodname{} but also serves as a stepping stone to creative generation, as generative AI progresses, continually expanding the limits of achievable creativity and artistic expression.

\section{Acknowledgements}

We extend our special thanks to Jia Wei Sii for her help in creating figures and discussing the main concept. We are also grateful to Ruoyi Du, Zhiyu Qu, Chee Seng Chan, and the reviewers for their fruitful comments and corrections on our draft, methodology, and experiments.


%% file: sec/X_suppl.tex
\clearpage
\appendix


\section{Implementation Details}

We conducted training on a single GeForce RTX 3090 GPU with a batch size of 2 over 100 epochs. AdamW \cite{loshchilov2018adamw} optimizer was employed with a constant learning rate of 0.0001 and weight decay of 0.01. Only random horizontal flip augmentation is used. $512 \times 512$ image resolution is applied.

We adopted the LoRA design \cite{hu2022lora} from \texttt{diffusers} library\footnote{\url{https://github.com/huggingface/diffusers/blob/main/examples/text_to_image/train_text_to_image_lora.py}}, in which the low-rank adapters were added to the $QKV$ and $out$ components of all cross-attention modules.

Regarding the attention loss (see Eq. (\textcolor{red}{5})), we selected cross-attention maps with a feature map size of $16 \times 16$. The specific layers chosen for this purpose were as follows:

{
\scriptsize
\begin{itemize}
    \item \texttt{down\_blocks.2.attentions.0.transformer\_blocks.0.attn2}
    \item \texttt{down\_blocks.2.attentions.1.transformer\_blocks.0.attn2}
    \item \texttt{up\_blocks.1.attentions.0.transformer\_blocks.0.attn2}
    \item \texttt{up\_blocks.1.attentions.1.transformer\_blocks.0.attn2}
    \item \texttt{up\_blocks.1.attentions.2.transformer\_blocks.0.attn2}
\end{itemize}
}

\section{Implementation of EMR and CoSim}

\begin{algorithm}[h]
\caption{EMR and CoSim for part composition}
\label{alg:emr_cosim_vcc}
\definecolor{codeblue}{rgb}{0.25,0.5,0.5}
\lstset{
  backgroundcolor=\color{white},
  basicstyle=\fontsize{7.2pt}{7.2pt}\ttfamily\selectfont,
  columns=fullflexible,
  breaklines=true,
  captionpos=b,
  commentstyle=\fontsize{7.2pt}{7.2pt}\color{gray},
  keywordstyle=\fontsize{7.2pt}{7.2pt}\color{codeblue},
}
\begin{lstlisting}[language=python]
# part_predictor: obtained via Sec 3.1
# pipeline: diffusion generation pipeline
# real_xs: real image (Nx512x512x3) where N up to 4
# M: number of parts
# D: number of dino feature dimension

# obtain the prompt of the real image (Eq.3)
p_input = part_predictor.<@\textcolor{violet}{\textbf{predict}}@>(real_xs[0])  # (M+1)
p_idxs = [0,1,...,M]
    
for real_x in real_xs[1:]:
    p_real = part_predictor.<@\textcolor{violet}{\textbf{predict}}@>(real_x)  # (M+1)

    # replace one part
    rand_idx = randint(len(p_idxs))
    rand_pop = p_idxs.pop(rand_idx)
    p_input[rand_pop] = p_real[rand_pop]
    
gen_x = <@\textcolor{violet}{\textbf{pipeline}}@>(p_input)
p_gen = part_predictor.<@\textcolor{violet}{\textbf{predict}}@>(gen_x)    # (M+1)

p_input_embs = part_predictor.<@\textcolor{violet}{\textbf{get\_centroids}}@>(p_input)  # (M+1,D)
p_gen_embs = part_predictor.<@\textcolor{violet}{\textbf{get\_centroids}}@>(p_gen)    # (M+1,D)

EMR = <@\textcolor{violet}{\textbf{average}}@>(p_input == p_gen)
CoSim = <@\textcolor{violet}{\textbf{average}}@>(<@\textcolor{violet}{\textbf{cossim}}@>(p_input_embs, p_gen_embs))
    
\end{lstlisting}
\end{algorithm}

\begin{algorithm}[h]
\caption{EMR and CoSim for part reconstruction}
\label{alg:emr_cosim}
\definecolor{codeblue}{rgb}{0.25,0.5,0.5}
\lstset{
  backgroundcolor=\color{white},
  basicstyle=\fontsize{7.2pt}{7.2pt}\ttfamily\selectfont,
  columns=fullflexible,
  breaklines=true,
  captionpos=b,
  commentstyle=\fontsize{7.2pt}{7.2pt}\color{gray},
  keywordstyle=\fontsize{7.2pt}{7.2pt}\color{codeblue},
}
\begin{lstlisting}[language=python]
# part_predictor: obtained via Sec 3.1
# pipeline: diffusion generation pipeline
# real_x: real image (512x512x3)
# M: number of parts
# D: number of dino feature dimension

# obtain the prompt of the real image (Eq.3)
p_real = part_predictor.<@\textcolor{violet}{\textbf{predict}}@>(real_x)  # (M+1)
gen_x = <@\textcolor{violet}{\textbf{pipeline}}@>(p_real)
p_gen = part_predictor.<@\textcolor{violet}{\textbf{predict}}@>(gen_x)    # (M+1)

# an example of "p" is [4, 222, 55, 23, 98, 22]
# in the "pipeline", we prepend word template like "a photo of a "
# e.g., "a photo of a [0,4] [1,222] ... [M,K]"
# the token [*,*] will be replaced by its embedding computed via Eq.4

p_real_embs = part_predictor.<@\textcolor{violet}{\textbf{get\_centroids}}@>(p_real)  # (M+1,D)
p_gen_embs = part_predictor.<@\textcolor{violet}{\textbf{get\_centroids}}@>(p_gen)    # (M+1,D)

EMR = <@\textcolor{violet}{\textbf{average}}@>(p_real == p_gen)
CoSim = <@\textcolor{violet}{\textbf{average}}@>(<@\textcolor{violet}{\textbf{cossim}}@>(p_real_embs, p_gen_embs))
    
\end{lstlisting}
\end{algorithm}

Our evaluation algorithms for the Exact Matching Rate (EMR) and Cosine Similarity (CoSim) between generated and real images are presented in Algorithms~\ref{alg:emr_cosim_vcc} and \ref{alg:emr_cosim}, respectively. Each algorithm is designed to evaluate a single sample. For the evaluations in Section \textcolor{red}{4.1}, 
we computed the average results over 500 iterations using Algorithm~\ref{alg:emr_cosim_vcc}. Similarly, for Section \textcolor{red}{4.2}, 
we averaged the outcomes over 5,994 and 12,000 iterations for the CUB-200-2011 (birds) and Stanford Dogs datasets, respectively.

\clearpage
\section{Examples of our part discovery}

\begin{figure}[h]
    \centering
    \includegraphics[width=\linewidth]{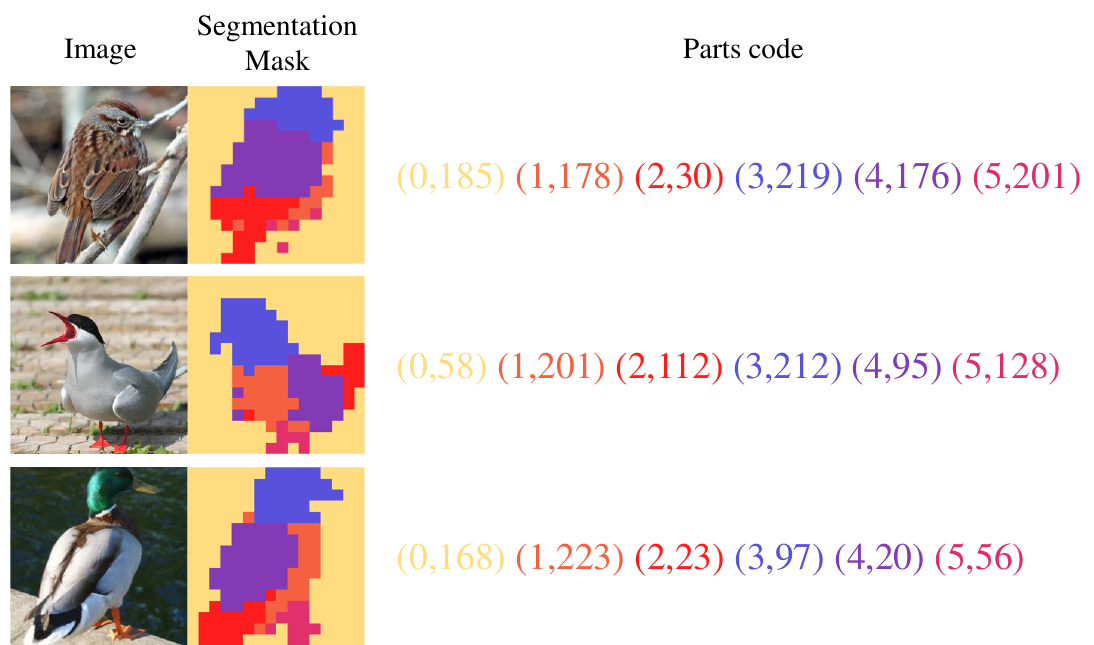}
    \caption{Three example outputs of our part discovery. Note that all these discrete IDs can be translated easily with minimal effort. For instance, 0 is background, 1 is tail, etc.}
    \label{fig:example_seg}
\end{figure}

In Fig.~\ref{fig:example_seg}, we display a few examples of our obtained segmentation masks and associated sets of parts.

\clearpage
\section{More examples}




\begin{figure}[h!]
    \centering
    \begin{subfigure}[b]{0.49\linewidth}
        \includegraphics[width=\linewidth]{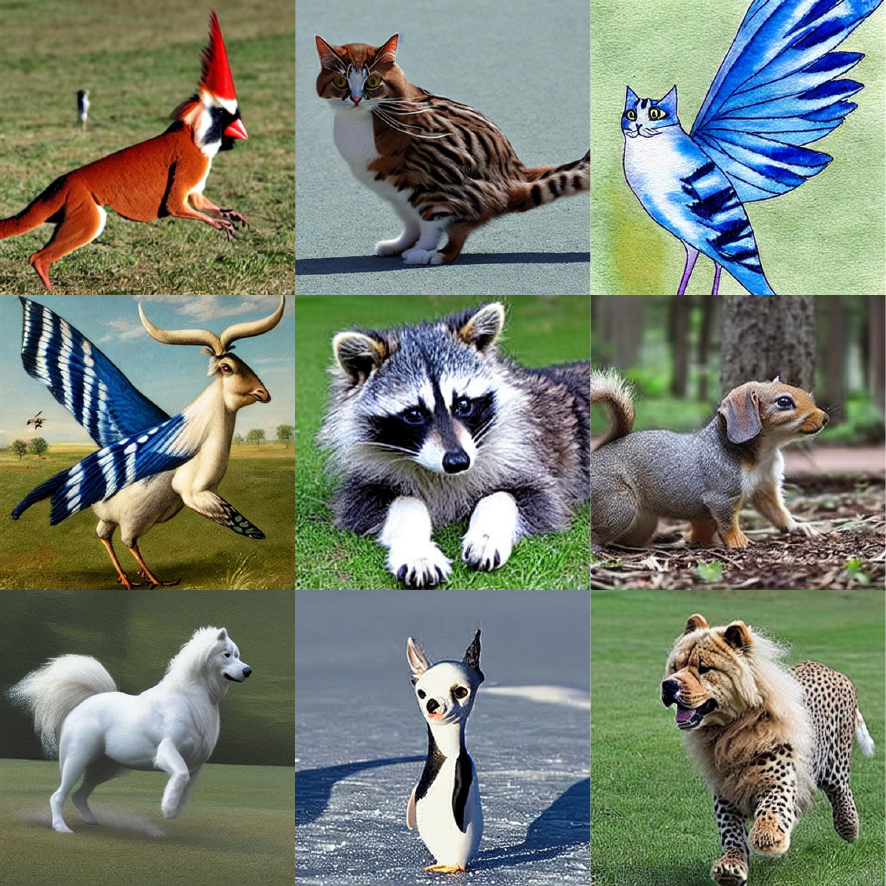}
        \caption{}
    \end{subfigure}
    \begin{subfigure}[b]{0.49\linewidth}
        \includegraphics[width=\linewidth]{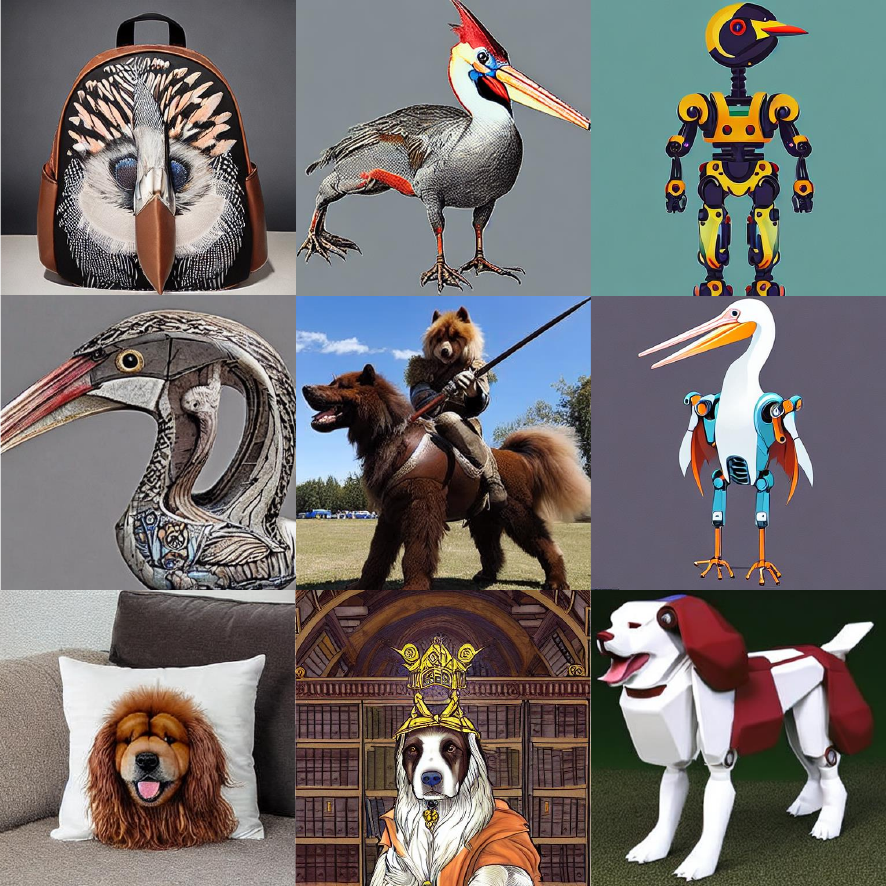}
        \caption{}
    \end{subfigure}

    \caption{We present additional examples of creative generation. \textbf{(a)} displays the effects of transferring learned parts, \eg, replacing a leopard head with a \texttt{chow}'s head. \textbf{(b)} displays using the learned parts to inspire some character/product designs.}
    \label{fig:design}
\end{figure}

\begin{figure}[h!]
    \centering
    \begin{subfigure}[b]{0.49\linewidth}
        \includegraphics[width=\linewidth]{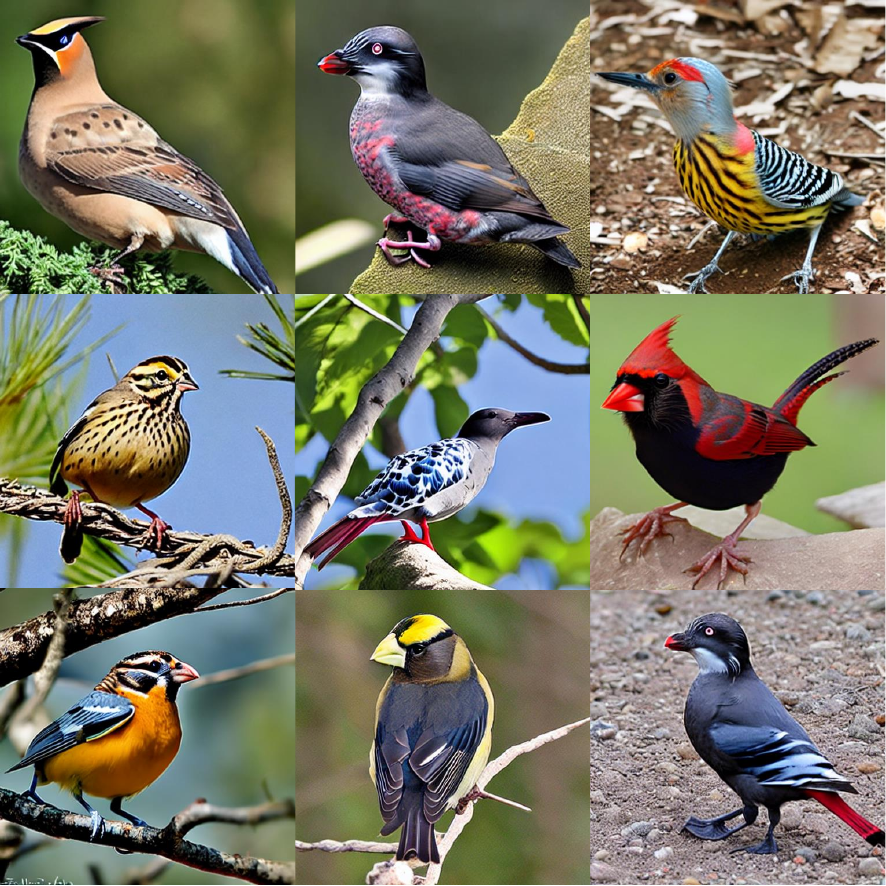}
        \caption{}
    \end{subfigure}
    \begin{subfigure}[b]{0.49\linewidth}
        \includegraphics[width=\linewidth]{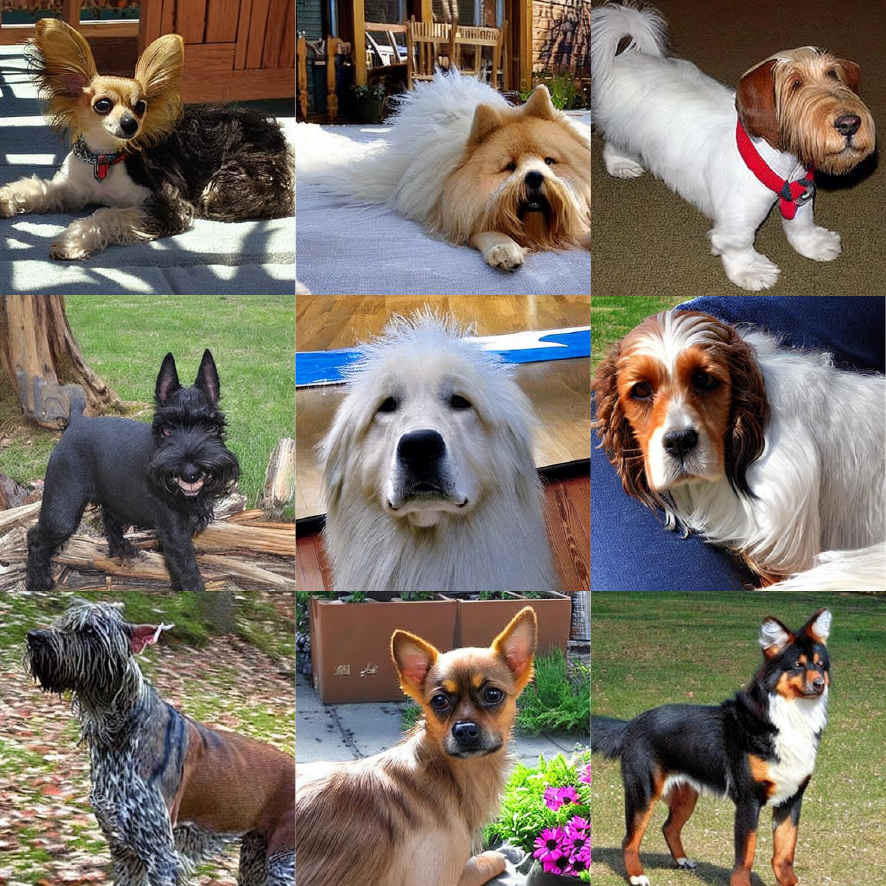}
        \caption{}
    \end{subfigure}

    \caption{We present additional examples of images generated by our PartCraft, featuring a random selection of parts.}
    \label{fig:uncurated_bird}
\end{figure}

    

\clearpage
\subsection{PartCraft for Character Face Creation}

\begin{figure}[h!]
    \centering
    \includegraphics[width=\linewidth]{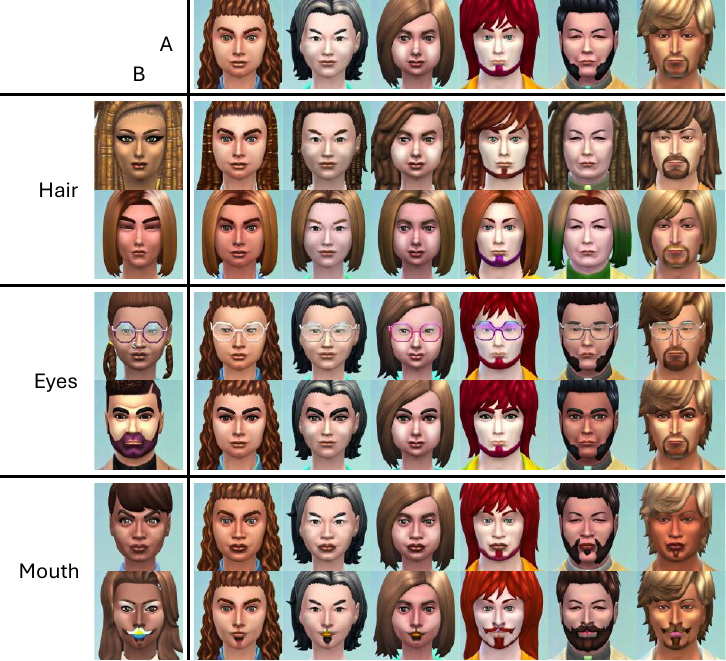}
    \caption{We present additional examples of images generated by our PartCraft, using data from Sims4-Faces \cite{sims4faces2022}. We transfer three different parts (\ie, hair, eyes, mouth) from source B to target A.}
    \label{fig:sims4}
\end{figure}

To further demonstrate the capability of PartCraft in learning parts, we conducted training on the publicly available Sims4-Faces \cite{sims4faces2022} dataset. We selected a subset of 200 faces (100 men and 100 women) from the dataset and applied our part discovery algorithm to parse the faces into various parts (\eg, hair, eyes, mouth, ear, neck). In Fig.~\ref{fig:sims4}, two sets of images were generated from their original parts (sources A and B), and a specific part from source B can be integrated into target A. This experiment shows that our PartCraft can work on domains where the creation and manipulation of parts are essential.



\clearpage
\section{Further Analysis}

\subsection{Attention loss weight}

\begin{figure}[h]
    \centering
    \includegraphics[scale=0.6]{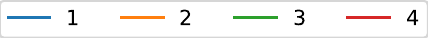} \\ 
    \includegraphics[scale=0.5]{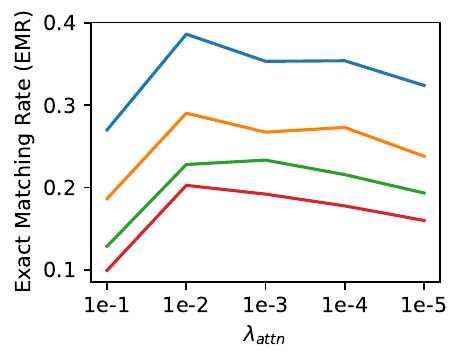}
    \includegraphics[scale=0.5]{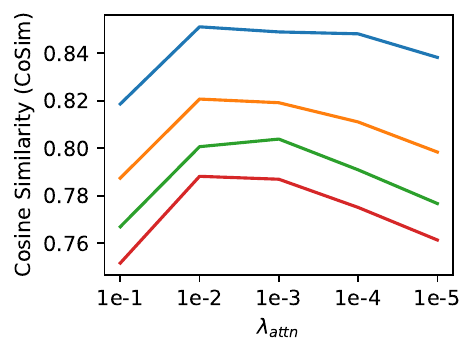}
    \caption{Ablation on the effect of $\lambda_{attn}$ for virtual creature generation on CUB-2011 birds. Different colors represent different numbers of composited parts.}
    \label{fig:ablattn_vcc}
\end{figure}

In Fig.~\ref{fig:ablattn_vcc} and Tab.~\ref{tab:ablattn}, we summarize the results of our ablation study on the impact of $\lambda_{attn}$. We observed that $\lambda_{attn}=0.01$ frequently yields the best EMR and CoSim scores, while also delivering comparable FID scores. Consequently, we have adopted this value as the default in our experiments.

\begin{table}[h]
    \centering
    \begin{tabular}{c|ccccc}
        $\lambda_{attn}$ & 0.1 & 0.01 & 0.001 & 0.0001 & 0.00001 \\
        \midrule
        FID ($\downarrow$) & 19.08 & 12.86 & 12.44 & 11.78 & \bf 11.64 \\
        EMR ($\uparrow$) & 0.339 & \bf 0.460 & 0.445 & 0.425 & 0.397 \\
        CoSim ($\uparrow$) & 0.851 & \bf 0.882 & 0.880 & 0.878 & 0.872 \\
    \end{tabular}
    \caption{Ablation on the effect of $\lambda_{attn}$ for conventional generation on CUB-200-2011 birds.}
    \label{tab:ablattn}
\end{table}

\subsection{Hyperparameter of Break-a-Scene}

We show the $\lambda=0.1/0.001$ result (CUB-200-2011). We hypothesize that stronger attention loss may cause overfitting and neglect the generation quality, thus lowering EMR/CoSim. Thus, we use $\lambda=0.01$ for Break-a-Scene experiments.

\begin{table}[h]
    \centering
    \adjustbox{max width=\linewidth}{
        \begin{tabular}{c|ccc}
            $\lambda$ & 0.1 & 0.01 & 0.001 \\
            \hline
            EMR & 0.366 & 0.390 & 0.344 \\
            CoSim & 0.841 & 0.854 & 0.832 \\
        \end{tabular}
    }
\end{table}

